# Belief and Truth in Hypothesised Behaviours

Stefano V. Albrecht[a], Jacob W. Crandall[b], Subramanian Ramamoorthy[c]

[a]*The University of Texas at Austin, United States*
[b]*Masdar Institute of Science and Technology, United Arab Emirates*
[c]*The University of Edinburgh, United Kingdom*

**Abstract**

There is a long history in game theory on the topic of Bayesian or "rational" learning, in which each player maintains beliefs over a set of alternative behaviours, or types, for the other players. This idea has gained increasing interest in the artificial intelligence (AI) community, where it is used as a method to control a single agent in a system composed of multiple agents with unknown behaviours. The idea is to hypothesise a set of types, each specifying a possible behaviour for the other agents, and to plan our own actions with respect to those types which we believe are most likely, given the observed actions of the agents. The game theory literature studies this idea primarily in the context of equilibrium attainment. In contrast, many AI applications have a focus on task completion and payoff maximisation. With this perspective in mind, we identify and address a spectrum of questions pertaining to belief and truth in hypothesised types. We formulate three basic ways to incorporate evidence into posterior beliefs and show when the resulting beliefs are correct, and when they may fail to be correct. Moreover, we demonstrate that prior beliefs can have a significant impact on our ability to maximise payoffs in the long-term, and that they can be computed automatically with consistent performance effects. Furthermore, we analyse the conditions under which we are able complete our task optimally, despite inaccuracies in the hypothesised types. Finally, we show how the correctness of hypothesised types can be ascertained during the interaction via an automated statistical analysis.

## 1. Introduction

There is a long history in game theory on the topic of Bayesian or "rational" learning (e.g. Nachbar, 2005; Dekel et al., 2004; Kalai and Lehrer, 1993; Jordan, 1991). Therein, players maintain beliefs about the behaviours, or "types", of other players in the form of a probability distribution over a set of alternative types. These beliefs are updated based on the observed actions, and each player chooses an action which is expected to maximise the payoffs received by the player, given the current beliefs of the player. The principal questions studied in this context are the degree to which players can learn to make correct predictions, and whether the interaction process converges to solutions such as Nash equilibrium (Nash, 1950).

This general idea, which we here refer to as the *type-based method*, has received increasing interest in the artificial intelligence (AI) community, where it is used as a method to control a single agent in a system composed of multiple agents (e.g. Albrecht and Ramamoorthy, 2013a; Barrett et al., 2011; Gmytrasiewicz and Doshi, 2005; Carmel and Markovitch, 1999). This interest is, in part, motivated by applications that require efficient and flexible interaction with agents whose behaviours are initially unknown. Example applications include adaptive user interfaces, robotic elderly care, and automated trading agents. Learning to interact from scratch in such settings is notoriously difficult, due to the essentially unconstrained nature of what the other agents may be doing and the fact that their behaviours are a priori unknown. The type-based method is seen as a way to reduce the complexity of such problems by focusing on a relatively small set of points in the infinite space of possible behaviours.

More concretely, the idea is to hypothesise ("guess") a set of types, each of which specifies a possible behaviour for the other agents. A type may be of any structural form, and here we simply view it as a "blackbox" programme which takes as input the interaction history and chooses actions for the next step in the interaction. Such types may be specified manually by a domain expert or generated automatically, e.g. from a corpus of historical data or the problem description. By comparing the predictions of the types with the observed actions of the agents, we can form posterior beliefs about the relative likelihood of types. The beliefs and types are in turn utilised in a planning procedure to find an action which maximises our expected payoffs with respect to our beliefs. A useful feature of this method is the fact that we may hypothesise any types of behaviours, which gives us the flexibility to interact with a variety of agents. Moreover, since each type specifies a complete behaviour, we can plan actions in the entire interaction space, including in situations that have not been encountered before.

Nonetheless, there are several questions and concerns associated with this method, pertaining to the evolution and impact of beliefs as well as the implications and detection of incorrect hypothesised types. Specifically, how should evidence (i.e. observed actions) be incorporated into beliefs and under what conditions will the beliefs be correct? What impact do prior beliefs have on our ability to maximise payoffs in the long-term? Furthermore, under what conditions will we be able to complete our task even if our hypothesised types are incorrect? And, finally, how can we ascertain the correctness of our hypothesised types during the interaction?

The AI literature on the type-based method has focused on experimental evaluations, exploration mechanisms, and computational issues arising from recursive beliefs, but not or only partially on the questions outlined above. (We defer a detailed discussion of related works to Section 2). On the other hand, the game theory literature addresses such questions primarily in the context of equilibrium attainment in repeated games (cf. Section 2). However, there are several reasons why this renders the game theory literature of limited applicability to domains such as the ones mentioned earlier. First, equilibrium concepts such as Nash equilibrium are based on normative assumptions, including perfect rationality with respect to one's payoffs. However, such normative assumptions are difficult to justify in situations in which we assume no prior knowledge about the behaviour of other agents. For example, there is evidence that humans do not satisfy such strict assumptions (e.g. Kahneman and Tversky, 1979). Second, an equilibrium solution prescribes behaviours for all involved agents, whereas we control only a single agent and assume no control over the choice of behaviour for the other agents. Finally, the existence of multiple equilibria with possibly differing payoff profiles means that equilibrium attainment may itself not be synonymous with payoff maximisation for our controlled agent.

The purpose of the present article is to improve our understanding of the type-based method by providing insight into the questions outlined above. Our analysis is based on stochastic Bayesian games, which are an extension of Bayesian games (Harsanyi, 1967) that include stochastic state transitions, and *Harsanyi-Bellman Ad Hoc Coordination* (HBA), which can be viewed as a general algorithmic description of the type-based method (Albrecht and Ramamoorthy, 2013a). After discussing related work in Section 2 and technical preliminaries in Section 3, the article makes the following contributions:

- **Section 4** considers three basic methods to incorporate observations into posterior beliefs and analyses the conditions under which they converge to the true distribution of types, including in processes in which type assignments may be randomised and correlated. We also discuss examples to show when beliefs may fail to converge to the correct distribution.

- **Section 5** investigates the impact of prior beliefs on payoff maximisation in a comprehensive empirical study. We show that prior beliefs can indeed have a significant impact on the long-term performance of HBA, and that the magnitude of the impact depends on the depth of the planning horizon (i.e. how far we look into the future). Moreover, we show that automatic methods can compute prior beliefs with consistent performance effects.

- **Section 6** analyses what relation the hypothesised types must have to the true types in order for HBA to be able to complete its task, despite inaccuracies in the hypothesised types. We formulate a hierarchy of increasingly desirable termination guarantees and analyse the conditions under which they are met. In particular, we give a novel characterisation of optimality which is based on the concept of probabilistic bisimulation (Larsen and Skou, 1991).

- **Section 7** shows how the truth of hypothesised types can be contemplated during the interaction in the form of an automated statistical analysis. The presented algorithm can incorporate multiple statistical features into the test statistic and learns its distribution during the interaction process, with asymptotic correctness guarantees. We show in a comprehensive set of experiments that the algorithm achieves high accuracy and scalability at low computational costs.

Finally, Section 8 concludes this work and discusses directions for future work. Elements of this work appeared in (Albrecht et al., 2015; Albrecht and Ramamoorthy, 2015, 2014, 2013b).

## 2. Related Work

This section discusses related work and situates our work within the literature. We distinguish between research on the type-based method in the areas of game theory and artificial intelligence.

### 2.1. Type-based Method in Game Theory

Perhaps the earliest formulation of the type-based method was in the form of *Bayesian games* (Harsanyi, 1967, 1968a,b). Bayesian games were introduced to address incomplete information games, in which certain aspects of the game are known to some players and unknown to others. Harsanyi proposed to model this "private information" as *types*: every player has one of a number of types which govern the player's behaviour[1], and the assignment of types is governed by some distribution over types. By assuming that the type spaces and distribution are common knowledge, this reduces the incomplete information game to a complete (but imperfect) information game, admitting a solution in the form of the Bayesian Nash equilibrium. While this idea was controversial at the time[2], Bayesian games have become a firm part of game theory.

---

[1]The interpretation of types as behaviours is consistent with the original definition of Harsanyi, who defines types as parameters for both payoff and strategy functions (cf. Section 7 in Harsanyi, 1967). See also Dekel et al. (2004).

[2]The controversy was centred around the assumption that players a priori know the true distribution of types. (From a personal conversation with Reinhard Selten.)

The model used in our work builds on Bayesian games but includes stochastic state transitions, making it more naturally applicable to many problems of interest in artificial intelligence. This also allows us to define concisely what it means to *complete* a task, namely to drive the game from an initial state into a terminal state. Moreover, in contrast to Bayesian games, we explicitly consider cases in which the type spaces and distribution are unknown to our agent, and we do not assume that other agents necessarily use a type-based reasoning or have common prior beliefs.

Much work in game theory has focused on equilibrium attainment as the result of *learning* through repeated interaction, in games in which players maintain Bayesian beliefs about the strategies of other players. In the seminal work of Kalai and Lehrer (1993), the authors show that under a certain assumption about players' beliefs called "absolute continuity" (essentially, every event that has true positive probability is assigned positive probability under the player's belief), players prediction of future play will become arbitrarily close to the true future play. A related result was shown by Jordan (1991) for myopic players which consider only immediate payoffs. In Section 4, we show that the convergence result of Kalai and Lehrer (1993) carries over to our model, and we also provide convergence results for different formulations of posterior beliefs which can recognise randomised and correlated type assignments.

In addition to posterior beliefs, it has been shown that prior beliefs are intimately connected to the equilibrium solution that emerges as a result of learning. For example, Nyarko (1998) use a similar but weaker condition than absolute continuity and show that the resulting subjective equilibrium may not be a Nash equilibrium if the players have different prior beliefs. Similarly, Dekel et al. (2004) show under certain conditions that learning without common prior beliefs may converge to a self-confirming equilibrium (Fudenberg and Levine, 1993) which is not a Nash equilibrium. While important in the context of equilibrium attainment, these results are less applicable to our focus on individual payoff maximisation and task completion (cf. Section 1). In Section 5, we show that prior beliefs can, nevertheless, have a significant impact on our ability to maximise payoffs in the long-term. Moreover, our results indicate that prior beliefs can be computed automatically with consistent performance effects.

The possibility of discrepancies between predicted and true behaviour has been recognised in works such as (Nachbar, 2005; Foster and Young, 2001; Nachbar, 1997). Essentially, these works show for certain games and conditions that players maintaining beliefs over behaviours cannot simultaneously make correct predictions and play optimally with respect to their beliefs. In Section 6, we consider the impact of incorrect hypothesised types on our ability to complete tasks and show that a certain form of optimality is preserved under a bisimulation relation, which can be verified in practice. Furthermore, in Section 7 we describe an automatic statistical analysis to allow an agent to contemplate the correctness of its behavioural hypotheses.

*2.2. Type-based Method in Artificial Intelligence*

There is a substantial body of work in the AI literature on coordination (e.g. Kaminka and Frenkel, 2007; Tambe, 1997; Grosz and Kraus, 1996) and learning (e.g. Conitzer and Sandholm, 2007; Hu and Wellman, 2003; Bowling and Veloso, 2002; Littman, 1994) in multiagent systems. However, it has been noted (e.g. Stone et al., 2010) that many of these methods depend on some form of prior coordination between agents. The type-based method has been studied as an alternative method of interaction with agents whose behaviours are initially unknown.

Barrett et al. (2011) implement a variant of the type-based method in the "pursuit" grid-world domain and demonstrate its practical potential. Albrecht and Ramamoorthy (2013a) introduce a general algorithm called *Harsanyi-Bellman Ad Hoc Coordination* (HBA) (cf. Section 3) and evaluate it in the "level-based foraging" grid-world domain and in matrix games played against humans. Both works propose various implementations of the type-based method, including tree expansion, dynamic programming, and reinforcement learning with stochastic sampling.

Carmel and Markovitch (1999) define types as deterministic finite state machines and study optimal exploration in repeated games. Similarly, Chalkiadakis and Boutilier (2003) use types in the context of multiagent reinforcement learning and develop exploration methods based on the concept of "value of information" (Howard, 1966). Their work is essentially an extension of Dearden et al. (1999), which study the related idea of maintaining Bayesian beliefs over a set of environment models in reinforcement learning.

Southey et al. (2005) apply the type-based method to variants of the poker game. The poker domain differs from the above works in that the state of the interaction process (i.e. player hands) is only partially observable. The authors show how beliefs can be maintained in this setting and compare various methods to compute optimal responses with respect to beliefs.

In interactive partially observable Markov decision processes (I-POMDPs) (Gmytrasiewicz and Doshi, 2005), agents make decisions in the presence of uncertainty regarding the state of the environment, the types of other agents, and their action choices. Several solution methods have been developed for I-POMDPs (e.g. Doshi et al., 2009; Doshi and Gmytrasiewicz, 2009; Doshi and Perez, 2008) and there have been attempts to apply I-POMDPs in practice (e.g. Doshi et al., 2010; Ng et al., 2010). An interesting parallel to our work is that the convergence result of Kalai and Lehrer (1993) has also been extended to I-POMDPs (Doshi and Gmytrasiewicz, 2006).

Bowling and McCracken (2005) use "play books" to control a single agent in a team of agents. Plays are similar to types but specify behaviours for a complete team and include additional structure such as applicability and termination conditions, and roles for each agent. Similarly, "plan libraries" have been used to infer an agent's goals (Carberry, 2001; Charniak and Goldman, 1993). Plans resemble types but may include intricate structure such as temporal and causal orderings, and grammars (Sukthankar et al., 2014; Geib and Goldman, 2009).

The above works investigate various aspects of the type-based method, but they do not or only partially address the questions outlined in Section 1. Specifically, most of the above works use a posterior formulation in which the likelihood is defined as a product of action probabilities. In Section 4, we show under what conditions this formulation will produce correct and *incorrect* beliefs, and we also investigate alternative posterior formulations. Moreover, only Chalkiadakis and Boutilier (2003) consider the effects of prior beliefs by comparing "uninformed" (i.e. uniform) and "informed" (uniform with narrowed support) prior beliefs, but they provide no detailed analysis. In Section 5, we investigate how prior beliefs affect our ability to maximise payoffs in the long-term and how they can be computed automatically. Finally, none of the above works consider the implications and detection of incorrect hypothesised types.

## 3. Model and Algorithm

This section introduces the general model and algorithm used in our work, and further elaborates on connections to other related works.

### 3.1. Stochastic Bayesian Game

We model the interaction process as a stochastic Bayesian game (SBG) (Albrecht and Ramamoorthy, 2013a) which can be viewed as a combination of the Bayesian game (Harsanyi, 1967) and the stochastic game (Shapley, 1953). This combination is useful because it allows us to study the type-based method of interaction via the established framework of Bayesian games while also providing a means to specify an environment (via states) and the task to be completed. The structural definition of SBGs is as follows:

**Definition 1.** A *stochastic Bayesian game* (SBG) consists of:

- finite state space $S$ with initial state $s^0 \in S$ and terminal states $\bar{S} \subset S$
- players $N = \{1, ..., n\}$ and for each $i \in N$:
  - finite set of actions $A_i$ (where $A = A_1 \times ... \times A_n$)
  - infinite type space $\Theta_i$ (where $\Theta = \Theta_1 \times ... \times \Theta_n$)
  - payoff function $u_i : S \times A \times \Theta_i \to \mathbb{R}$
  - strategy function $\pi_i : \mathbb{H} \times A_i \times \Theta_i \to [0, 1]$
- state transition function $T : S \times A \times S \to [0, 1]$
- type distribution $\Upsilon : \Theta^+ \to [0, 1]$, where $\Theta^+$ is a finite subset of $\Theta$

and $\mathbb{H}$ denotes the set of all *histories* $H^t = \langle s^0, a^0, s^1, a^1, ..., s^t \rangle$ with $t \geq 0$, such that $s^0, ..., s^t \in S$ and $a^0, ..., a^{t-1} \in A$.

A SBG defines the interaction process as follows:

**Definition 2.** A SBG starts at time $t = 0$ in state $s^0$:

1. In state $s^t$, the types $\theta_1^t, ..., \theta_n^t$ are sampled from $\Theta^+$ with probability $\Upsilon(\theta_1^t, ..., \theta_n^t)$, and each player $i$ is informed only about its own type $\theta_i^t$.

2. Based on the history $H^t$, each player $i$ chooses an action $a_i^t \in A_i$ with probability $\pi_i(H^t, a_i^t, \theta_i^t)$, resulting in the joint action $a^t = (a_1^t, ..., a_n^t)$.

3. Each player $i$ receives an individual payoff given by $u_i(s^t, a^t, \theta_i^t)$, and the game transitions into a successor state $s^{t+1} \in S$ with probability $T(s^t, a^t, s^{t+1})$.

This process is repeated until a terminal state $s^t \in \bar{S}$ is reached, after which the game stops.

Throughout this work, we will use the contextual notation $H^\tau$ to denote the $\tau$-prefix of $H^t$ (i.e. $H^\tau$ is the initial segment of $H^t$ up until state $s^\tau$, with $\tau \leq t$). Similarly, we use $s^\tau$ and $a^\tau$ to denote the respective $\tau$-elements of $H^t$.

The set $S$ can be used to specify the *environment* within which the players interact, where each state $s \in S$ is a specific configuration of the environment. For instance, the environment may be a maze in a two-dimensional grid and the states may specify the positions of players and walls. The task in the SBG is to drive the interaction process from the initial state to a terminal state. Once a terminal state is reached, we say that the *task is completed*.

The type space $\Theta_i$ contains all possible behaviours for player $i$. Each type $\theta_i \in \Theta_i$ corresponds to a complete behaviour for player $i$ by specifying its preferences, via $u_i$, and the way in which it chooses actions, via $\pi_i$ (see also Footnote 1). We place no restrictions on the behaviours that players can exhibit; in particular, each player can make decisions based on the entire history $H^t$. This includes behaviours that learn and change over time. In practice, it is useful to view a type as a *blackbox programme* which, through $\pi_i$, takes as input the current interaction history and returns probabilities for each action available to the player. (Sections 4 and 5 provide various examples of types; see also (Albrecht and Ramamoorthy, 2013a) for examples of types in complex SBGs.)

The types are assigned during the game via the type distribution $\Upsilon$. In this work, we consider two classes of types distributions:

**Definition 3.** A type distribution $\Upsilon$ is called *pure* if $\exists \theta \in \Theta^+ : \Upsilon(\theta) = 1$. A type distribution which is not pure is called *mixed*.

Pure type distributions specify one fixed type for each player, throughout the game. This is what we would normally expect, since it means that each player has a single coherent behaviour. However, there are cases in which it may make sense to assume a mixed type distribution. For example, Albrecht and Ramamoorthy (2013a) used a mixed type distribution in their human-machine experiments to allow for the possibility that human subjects may change between several simple types (as opposed to defining one complex type which includes the simple types).

Note that the type space $\Theta_i$ is uncountable because the strategy $\pi_i$ assigns *probabilities* to actions, and the interval $[0, 1]$ is itself uncountable. Therefore, in order for $\Upsilon$ to be a well-defined probability distribution, we define it over a finite or countable subset $\Theta^+ \subset \Theta$. (Otherwise, $\Upsilon$ would need to be defined as a density.) To differentiate the two spaces, we sometimes refer to $\Theta_i$ as the full type space and to $\Theta_i^+$ as the *true* types of player $i$. For convenience (and by abuse of notation), we will allow $\Upsilon(\theta)$ for any $\theta \in \Theta$, with $\Upsilon(\theta) = 0$ if $\theta \notin \Theta^+$.

### 3.2. Harsanyi-Bellman Ad Hoc Coordination

As outlined in Section 1, we consider a single agent which employs the type-based method to interact with other agents with unknown behaviours. Throughout this work, we use *Harsanyi-Bellman Ad Hoc Coordination* (HBA) (Albrecht and Ramamoorthy, 2013a) as a general algorithmic description of the type-based method. Algorithm 1 provides a formal definition of HBA.

Given a SBG $\Gamma$, we use $i$ to denote our player and $j$ and $-i$ to denote the other players (such as in $A_{-i} = \times_{j \neq i} A_j$). The behaviour of player $i$ is completely specified by HBA. In other words, $i$ has a single fixed type, $\Theta_i^+ = \{\theta_i^{\text{HBA}}\}$, where $\theta_i^{\text{HBA}}$ is defined by Algorithm 1. Thus, we may omit $\theta_i^{\text{HBA}}$ in $u_i$ and $\pi_i$ for compactness. The behaviour of the other players is governed by $\Upsilon$ and a priori unknown to us. Formally, we assume that all elements of $\Gamma$ are known to us except for $\Theta_j^+$ and $\Upsilon$, which are latent elements.

In HBA, these latent elements are essentially substituted for by the hypothesised type space $\Theta_j^*$ and the posterior belief Pr, respectively. Like $\Theta_j^+$, $\Theta_j^*$ is a finite or countable subset of the full type space $\Theta_j$. The posterior belief (probability) $\Pr(\theta_{-i}^* | H^t)$ quantifies the relative likelihood that players $j \neq i$ are of types $\theta_{-i}^* = (\theta_1^*, ..., \theta_{i-1}^*, \theta_{i+1}^*, ..., \theta_n^*)$, given the history $H^t$. If we assume independence of types, we can define Pr as

$$\Pr(\theta_{-i}^* | H^t) = \prod_{j \neq i} \Pr_j(\theta_j^* | H^t) \tag{1}$$

$$\Pr_j(\theta_j^* | H^t) = \frac{L(H^t | \theta_j^*) \, P_j(\theta_j^*)}{\sum_{\hat{\theta}_j^* \in \Theta_j^*} L(H^t | \hat{\theta}_j^*) \, P_j(\hat{\theta}_j^*)} \tag{2}$$

where $P_j(\theta_j^*)$ is the prior belief (probability) that player $j$ is of type $\theta_j^*$ *before* any actions are observed, and $L(H^t | \theta_j^*)$ is the (non-negative) likelihood of history $H^t$ assuming that player $j$ is of type $\theta_j^*$. It is convenient to define $\Pr_j(\theta_j^* | H^0) = P_j(\theta_j^*)$. Note that the likelihood $L$ in (2) is unspecified at this point; we will consider two variants for $L$ in Section 4.

The independence assumption of types is prevalent in the works discussed in Section 2. In the game theory literature (cf. Section 2.1), it is justified by the fact that the Nash equilibrium assumes that players choose actions independently. (This is opposed to concepts such as correlated equilibrium (Aumann, 1974) in which action choices may be correlated.) From a practical perspective,

---
**Algorithm 1** Harsanyi-Bellman Ad Hoc Coordination (HBA)
---
**Input:** current history $H^t = \langle s^0, a^0, s^1, a^1, ..., s^t \rangle$

**Output:** action probabilities $\pi_i(H^t, a_i)$ for player $i$

**Parameters:** hypothesised type spaces $\Theta_j^*$ for players $j \neq i$, discount factor $\gamma \in [0, 1]$

*1.* For each $a_i \in A_i$, compute expected payoff $E_{s^t}^{a_i}(H^t)$ with

$$E_s^{a_i}(\hat{H}) = \sum_{\theta_{-i}^* \in \Theta_{-i}^*} \Pr(\theta_{-i}^* | \hat{H}) \sum_{a_{-i} \in A_{-i}} Q_s^{(a_i, a_{-i})}(\hat{H}) \prod_{j \neq i} \pi_j(\hat{H}, a_j, \theta_j^*) \quad (3)$$

$$Q_s^a(\hat{H}) = \sum_{s' \in S} T(s, a, s') \left[ u_i(s, a) + \gamma \max_{a_i \in A_i} E_{s'}^{a_i}\left(\langle \hat{H}, a, s' \rangle\right) \right] \quad (4)$$

*2.* Distribute $\pi_i(H^t, \cdot)$ uniformly over $\arg\max_{a_i \in A_i} E_{s^t}^{a_i}(H^t)$

---

another justification is the fact that, while types are assumed to be independent, the behaviours they encode may very well depend on the behaviour of other players. This is since each player can make decisions based on the entire interaction history, which includes the observed actions of other players. Nonetheless, Section 4 also discusses the possibility of *correlated* types.

Where do the hypothesised types $\theta_j^* \in \Theta_j^*$ come from? In this work, we assume that the user has some means to generate such hypotheses. One way is to have them specified manually by domain experts, based on their experience with the problem (e.g. Albrecht and Ramamoorthy, 2013a). Another method is to generate types automatically from the problem description. For example, in Sections 5 and 7 we use three different methods to automatically generate sets of types for any given matrix game. Finally, one may use machine learning methods to extract types from a corpus of historical data (e.g. Barrett et al., 2013; Gal et al., 2004).

HBA performs a planning procedure, defined by (3)/(4), to find an action which maximises its expected long-term payoff with respect to its current beliefs and hypothesised types. Formally, (3) corresponds to player $i$'s component of the Bayesian Nash equilibrium (Harsanyi, 1968a) and (4) corresponds to the Bellman optimality equation (Bellman, 1957). Intuitively, (3)/(4) expand a tree of all possible future trajectories of the interaction process and weight each trajectory based on the posterior beliefs and predicted action probabilities of the hypothesised types. Note that $H^t$ denotes the current history while $\hat{H}$ is used to construct all future trajectories (histories), where the notation $\langle \hat{H}, a, s' \rangle$ in (4) denotes concatenation of $\hat{H}$ and $(a, s')$.

In practice, HBA may be implemented by limiting the recursion in (3)/(4) to some fixed depth (e.g. as in Section 5). However, it is easy to see that this procedure has time complexity which is exponential in factors such as the number of players, actions, and states in the game. This can make it a very costly operation and usually requires more sophisticated *approximate* methods when applied to complex domains. In this regard, a promising approach is given by stochastic sampling methods such as those used in (Albrecht and Ramamoorthy, 2013a; Barrett et al., 2011). In this work, unless stated otherwise, we assume that (3)/(4) are implemented as given.

It is worth noting that HBA does not require explicit exploration methods (i.e. deliberately choosing actions which do not maximise $E_{s^t}^{a_i}(H^t)$) because exploration is *implicit* in the calculation of $E_{s^t}^{a_i}(H^t)$. Specifically, for each action $a_i$, $E_{s^t}^{a_i}(H^t)$ predicts the impact of $a_i$ on HBA's beliefs and future interaction. This allows HBA to reason about the benefit of choosing a particular action, in the sense of what information that action can potentially reveal to HBA (Howard, 1966). Of course, this assumes that the true types of other players are included in the hypothesised types. Nonetheless, when the predictive ability of HBA is limited (e.g. due to a fixed recursion depth; cf. Section 5) or if we use opponent modelling to learn new types during the interaction (e.g. Albrecht and Ramamoorthy, 2013a; Barrett et al., 2011), then it may still be worthwhile to use explicit exploration methods such as those discussed in (Carmel and Markovitch, 1999) or approximations as in (Chalkiadakis and Boutilier, 2003).

*3.3. Relation to Other Interactive Decision Models*

Section 2 provided an overview of related works and models used therein. Here, we further elaborate on the connections and differences to some of these models and other models.

As pointed out earlier, our SBG model can be viewed as a combination of Bayesian games and stochastic games. If we remove states from the definition of SBGs (or, equivalently, assume a single state and no terminal states), then this reduces to a standard Bayesian game. In this case, $\Theta_j^+$ corresponds to the type spaces used in Bayesian games and $\Upsilon$ corresponds to the "basic probability distribution" (Harsanyi, 1967). (However, note that in contrast to Bayesian games, we assume no knowledge of $\Theta_j^+$ and $\Upsilon$.) On the other hand, if we remove types from the definition of SBGs, then the model reduces to a standard stochastic game. However, note that Shapley (1953) considers Markovian ("stationary") strategies whereas we allow strategies to depend on the entire interaction history. This definition of strategies is consistent with the model used by Kalai and Lehrer (1993), in which strategies are mappings from histories to probability distributions over actions.

A central assumption in SBGs is that the states and chosen actions are fully observable by the players. This is in contrast to I-POMDPs (cf. Section 2) in which states and actions are not directly observed. Instead, players receive noisy and possibly incomplete signals that depend on the state, based on which players infer beliefs over states. This makes I-POMDPs a very general model, but it also increases their computational complexity significantly. Another difference is that SBGs allow for mixed type distributions while I-POMDPs generally assume fixed types. These differences mean that the results of our work may not directly carry over to I-POMDPs.

Other models of interactive decision making exist, such as the decentralised POMDP (Bernstein et al., 2000) and partially observable stochastic game (e.g. Emery-Montemerlo et al., 2004). Both of these models allow for partial observability of process states as described above. While these models do not explicitly encode types, it is possible to emulate types by using factored states which are composed of individual elements.[3] Essentially, we can define the factored state space $\hat{S} = S \times \Theta_1^+ \times ... \times \Theta_n^+$, where the $S$-element is observed by all players and controlled by their joint actions while the $\Theta_i^+$-elements are privately observed by the players and controlled by the type distribution. An interesting question, then, is to what extent solving this model may produce a similar or better solution than HBA. However, as we will discuss next, this leads to another crucial difference between our work and the above works.

Once a model is fully specified, the usual goal is to *solve* it via some procedure. In the context of game theory, a solution may be a profile of strategies that satisfy some equilibrium property (e.g. Etessami and Yannakakis, 2010; Conitzer and Sandholm, 2008). In the context of artificial intelligence, a solution is a control policy for one or more agents which satisfies certain guarantees such as payoff maximisation (e.g. Dibangoye et al., 2013; Doshi et al., 2009; Hansen et al., 2004).

---

[3]We thank an anonymous reviewer for suggesting this line of thought.

This is in contrast to our work, in which we do not attempt to solve SBGs in this sense. Instead, we *prescribe* a specific normative solution for a single agent, in the form of HBA. This is similar in spirit to works such as (Kalai and Lehrer, 1993), except that we only consider a single agent that uses HBA. The advantage of this approach is that HBA can be applied "instantly", without the need to solve the model beforehand. This means that HBA may be applied to problems which are too complex to be solved in the conventional sense. Of course, the disadvantage is that we do not exactly know how HBA will perform, and the purpose of the present work is precisely to provide answers to this question.

## 4. Correctness of Posterior Beliefs

A central aspect of the type-based method are the beliefs over types. Beginning with some initial beliefs about the relative likelihood of types, we compare the predictions of types with the observed actions and update our beliefs to reflect the given evidence. Associated with this process are two key questions: how may evidence be incorporated into beliefs, and under what conditions will the beliefs be correct? As can be seen in Algorithm 1, these are important questions since the accuracy of the expected payoffs (3) depends on the accuracy of the posterior belief Pr.

In this section, we consider three classes of type distributions to cover a broad spectrum of scenarios: pure distributions, in which all agents have a fixed type; mixed distributions, in which types are randomly re-allocated; and correlated distributions, in which type assignments may be correlated. Corresponding to these classes, we consider three formulations of posterior beliefs which prescribe different ways to incorporate evidence into beliefs. We provide theoretical conditions under which these formulations produce *correct* beliefs, and we provide examples to show when they may fail to do so.

Our definition of correctness is with respect to the type distribution $\Upsilon$: beliefs are said to be correct if they assign the same probabilities to true types as $\Upsilon$. This requires that the beliefs can point to the types in the support of the type distribution. Therefore, the results in this section pertain to a situation in which the user knows that the true type space $\Theta_j^+$ must be a subset of the hypothesised type space $\Theta_j^*$. Formally, we assume:

**Assumption 1.** $\forall j \neq i : \Theta_j^+ \subseteq \Theta_j^*$

The case in which beliefs cannot be correct as defined above, due to incomplete or incorrect hypothesised types, is examined in Sections 6 and 7.

Finally, recall from Section 3.2 that posterior beliefs of the form (1) assume independence of player types. That is, they assume that the type distribution $\Upsilon$ can be represented as a product of $n$ independent factors $\Upsilon_j$ (one for each player), such that $\Upsilon(\theta) = \prod_j \Upsilon_j(\theta_j)$. Hence, in the following, unless states otherwise, we assume that $\Upsilon$ satisfies this independence property. Section 4.3 also considers the case of correlated type distributions.

### 4.1. Product Posterior

We begin our analysis with the product posterior:

**Definition 4.** The *product posterior* is defined as (1) with

$$L(H^t|\theta_j^*) = \prod_{\tau=0}^{t-1} \pi_j(H^\tau, a_j^\tau, \theta_j^*). \tag{5}$$

This is the standard posterior formulation used in Bayesian games and most of the works discussed in Section 2. It can be shown that, under a pure type distribution and if HBA does not a priori rule out any of the types in $\Theta_j^*$, then it will learn to make correct future predictions. Let $H^\infty$ be an infinite history with prefix $H^\tau$, and denote by $P_\Upsilon(H^\tau, H^\infty)$ and $P_{\Pr}(H^\tau, H^\infty)$, respectively, the *true* probability (based on $\Upsilon$) and the probability assigned by HBA (based on Pr) that $H^\tau$ will continue as prescribed by $H^\infty$.

**Theorem 1.** Let $\Gamma$ be a SBG with a pure type distribution $\Upsilon$. If HBA uses a product posterior and if the prior beliefs $P_j$ are positive (i.e. $\forall \theta_j^* \in \Theta_j^* : P_j(\theta_j^*) > 0$), then:
for any $\epsilon > 0$, there is a time $t$ from which ($\tau \geq t$)

$$P_{\Pr}(H^\tau, H^\infty)(1 - \epsilon) \leq P_\Upsilon(H^\tau, H^\infty) \leq P_{\Pr}(H^\tau, H^\infty)(1 + \epsilon) \qquad (6)$$

for all $H^\infty$ with $P_\Upsilon(H^\tau, H^\infty) > 0$.

*Proof.* The proof is not difficult but tedious, hence we defer it to Appendix A. Proof sketch: Kalai and Lehrer (1993) studied a model which can be equivalently described as a single-state SBG ($|S| = 1$) with pure type distribution $\Upsilon$ and proved Theorem 1 within their model. Their convergence result can be extended to multi-state SBGs by translating the multi-state SBG $\Gamma$ into a single-state SBG $\hat{\Gamma}$ which is equivalent to $\Gamma$ in the sense that the players behave identically. Essentially, the trick is to remove the states in $\Gamma$ by introducing a new player whose action choices correspond to the state transitions in $\Gamma$. □

Theorem 1 states that HBA will eventually make correct future predictions when using a product posterior against a pure type distribution (assuming the prior beliefs are positive). However, there is a subtle but important asymmetry between making correct future predictions and knowing the true type distribution: while the latter implies the former, the reverse is not generally true. The following example[4] illustrates this:

**Example 1.** Consider a SBG with two players and two actions, C and D. Player 1 is controlled by HBA using a product posterior while player 2 has two types, $\Theta_2^+ = \{\theta_{\lambda=0.1}, \theta_{\lambda=0.5}\}$, which are assigned by some pure type distribution. The two types choose action $C$ if player 1 chose C in the previous round. Otherwise, with probability $\lambda$, they will forever play action D. In this case, HBA will never know the correct type with absolute certainty. Even if HBA chooses D and player 2 responds by playing D indefinitely, there is still no certainty because $\lambda > 0$ in both types.

Therefore, while HBA is guaranteed to make correct future predictions after some time, it is not guaranteed to learn the type distribution of the game. Finally, note that Theorem 1 pertains to pure type distributions only. The following example shows that the product posterior may fail in SBGs with mixed type distributions:

**Example 2.** Consider a SBG with two players. Player 1 is controlled by HBA using a product posterior while player 2 has two types, $\Theta_2^+ = \{\theta_A, \theta_B\}$, which are assigned by a mixed type distribution $\Upsilon$ with $\Upsilon(\theta_A) = \Upsilon(\theta_B) = 0.5$. The type $\theta_A$ always chooses action A while $\theta_B$ always chooses action B. In this case, there will be a time $t$ after which both types have been assigned at least once, and so both actions A and B have been played at least once by player 2. This means that from time $t$ and all subsequent times $\tau \geq t$, we have $\Pr_2(\theta_A | H^\tau) = \Pr_2(\theta_B | H^\tau) = 0$ (that is, $\Pr_2$ is undefined), and HBA will fail to make correct future predictions.

---

[4]All examples in this section assume $\Theta_j^* = \Theta_j^+$ and uniform prior beliefs $P_j(\theta_j^*) = |\Theta_j^*|^{-1}$.

*4.2. Sum Posterior*

We continue our analysis with the sum posterior:

**Definition 5.** The *sum posterior* is defined as (1) with

$$L(H^t|\theta_j^*) = \sum_{\tau=0}^{t-1} \pi_j(H^\tau, a_j^\tau, \theta_j^*). \tag{7}$$

The sum posterior allows HBA to recognise changing types. In other words, the purpose of the sum posterior is to learn mixed type distributions. It is easy to see that a sum posterior would indeed learn the mixed type distribution in Example 2. However, we now give an example to show that, without additional requirements, the sum posterior does not necessarily learn any (pure or mixed) type distribution:

**Example 3.** Consider a SBG with two players. Player 1 is controlled by HBA using a sum posterior while player 2 has two types, $\Theta_2^+ = \{\theta_A, \theta_{AB}\}$, which are assigned by a pure type distribution $\Upsilon$ with $\Upsilon(\theta_A) = 1$. The type $\theta_A$ always chooses action A while $\theta_{AB}$ chooses actions A and B with equal probability. While the product posterior converges to the correct probabilities $\Upsilon$, the sum posterior converges to probabilities $\langle \frac{2}{3}, \frac{1}{3} \rangle$, which is incorrect.

Note that this example can be readily modified to use a mixed type distribution, with similar results. Therefore, we conclude that, without further assumptions, the sum posterior does not necessarily learn any type distribution.

Under what condition is the sum posterior guaranteed to learn the true type distribution of the game? Consider the following two quantities, which can be computed from a given history $H^t$:

**Definition 6.** The *average overlap* of player $j$ in $H^t$ is defined as

$$\text{AO}_j(H^t) = \frac{1}{t} \sum_{\tau=0}^{t-1} \left[|\Lambda_j^\tau| \geq 2\right]_1 \sum_{\theta_j^* \in \Theta_j^*} \pi_j(H^\tau, a_j^\tau, \theta_j^*) |\Theta_j^*|^{-1} \tag{8}$$

$$\Lambda_j^\tau = \left\{\theta_j^* \in \Theta_j^* \mid \pi_j(H^\tau, a_j^\tau, \theta_j^*) > 0\right\} \tag{9}$$

where $[b]_1 = 1$ if $b$ is true, else 0.

**Definition 7.** The *average stochasticity* of player $j$ in $H^t$ is defined as

$$\text{AS}_j(H^t) = \frac{1}{t} \sum_{\tau=0}^{t-1} |\Theta_j^*|^{-1} \sum_{\theta_j^* \in \Theta_j^*} \frac{1 - \pi_j(H^\tau, \hat{a}_j^\tau, \theta_j^*)}{1 - |A_j|^{-1}} \tag{10}$$

where $\hat{a}_j^\tau \in \arg\max_{a_j} \pi_j(H^\tau, a_j, \theta_j^*)$.

Both quantities are bounded by 0 and 1. The average overlap describes the similarity of the types, where $\text{AO}_j(H^t) = 0$ means that player $j$'s types (on average) never chose the same action in history $H^t$, whereas $\text{AO}_j(H^t) = 1$ means that they behaved identically. The average stochasticity describes the uncertainty of the types, where $\text{AS}_j(H^t) = 0$ means that player $j$'s types (on average) were fully deterministic in the action choices in history $H^t$, whereas $\text{AS}_j(H^t) = 1$ means that they chose actions uniformly randomly.

**Example 4.** Consider the SBG from Example 3. Here, player 2 always chooses action A, since its type is always $\theta_A$. Therefore, for any history $H^t$, we have $\text{AO}_2(H^t) = 0.75$, which indicates a substantial amount of overlap between $\theta_A$ and $\theta_{AB}$. Furthermore, we have $\text{AS}_2(H^t) = 0.5$, which indicates a certain degree of randomisation. In fact, $\theta_A$ is fully deterministic while $\theta_{AB}$ is uniformly random, hence the average stochasticity in in the centre of the spectrum $[0, 1]$.

It can be shown that, if the average overlap and stochasticity of player $j$ converge to zero as $t \to \infty$, then the sum posterior is guaranteed to converge to any pure or mixed type distribution:

**Theorem 2.** Let $\Gamma$ be a SBG with a pure or mixed type distribution $\Upsilon$. If HBA uses a sum posterior, then, for $t \to \infty$: If $\text{AO}_j(H^t) = 0$ and $\text{AS}_j(H^t) = 0$ for all players $j \neq i$, then $\Pr(\theta_{-i}|H^t) = \Upsilon(\theta_{-i})$ for all $\theta_{-i} \in \Theta^+_{-i}$.

*Proof.* Throughout this proof, let $t \to \infty$. The sum posterior is defined as (1) where $L$ is defined as (7). Given the definition of $L$, both the numerator and the denominator in (2) may be infinite. We invoke L'Hôpital's rule which states that, in such cases, the quotient $\frac{u(t)}{v(t)}$ is equal to the quotient $\frac{u'(t)}{v'(t)}$ of the respective derivatives with respect to $t$. The derivative of $L$ with respect to $t$ is the average growth per time step, which in general may depend on the history $H^t$ of states and actions. The average growth of $L$ is

$$L'(H^t|\theta_j) = \sum_{a_j \in A_j} F(a_j|H^t) \pi_j(H^t, a_j, \theta_j) \tag{11}$$

where

$$F(a_j|H^t) = \sum_{\theta_j \in \Theta^+_j} \Upsilon(\theta_j) \pi_j(H^t, a_j, \theta_j) \tag{12}$$

is the probability of action $a_j$ after history $H^t$, with $\Upsilon(\theta_j)$ being the marginal probability that player $j$ is assigned type $\theta_j$. As we will see shortly, we can make an asymptotic growth prediction irrespective of $H^t$. Given that $\text{AO}_j(H^t) = 0$, we can infer that whenever $\pi_j(H^t, a_j, \theta^*_j) > 0$ for action $a_j$ and type $\theta^*_j$, then $\pi_j(H^t, a_j, \hat{\theta}^*_j) = 0$ for all other types $\hat{\theta}^*_j \neq \theta^*_j$ with $\hat{\theta}^*_j \in \Theta^*_j$. Therefore, we can write (11) as

$$L'(H^t|\theta_j) = \Upsilon(\theta_j) \sum_{a_j \in A_j} \pi_j(H^t, a_j, \theta_j)^2 \tag{13}$$

Next, given that $\text{AS}_j(H^t) = 0$, we know that there exists an action $a_j$ in (13) with $\pi_j(H^t, a_j, \theta_j) = 1$, and, therefore, we can conclude that $L'(H^t|\theta_j) = \Upsilon(\theta_j)$. This shows that the history $H^t$ is irrelevant to the asymptotic growth rate of $L$. Finally, since $\sum_{\theta_j \in \Theta^+_j} \Upsilon(\theta_j) = 1$, we know that the denominator in (2) will be 1, and we conclude that $\Pr_j(\theta_j|H^t) = \Upsilon(\theta_j)$. □

Theorem 2 explains why the sum posterior converges to the correct type distribution in Example 2. Since the types $\theta_A$ and $\theta_B$ always choose different actions and are completely deterministic (i.e. the average overlap and stochasticity are always zero), the sum posterior is guaranteed to converge to the type distribution. On the other hand, in Example 3 the types $\theta_A$ and $\theta_{AB}$ produce an overlap whenever action A is chosen, and $\theta_{AB}$ is completely random. Therefore, the average overlap and stochasticity are always positive, and an incorrect type distribution was learned.

The assumptions made in Theorem 2, namely that the average overlap and stochasticity converge to zero, require practical justification. First of all, it is important to note that it is only required that these converge to zero *on average* as $t \to \infty$. This means that in the beginning there may be arbitrary overlap and stochasticity, as long as these go to zero as the game proceeds. In fact, with respect to stochasticity, this is precisely how the exploration-exploitation dilemma (Sutton and Barto, 1998) is solved in practice: In the early stages, the agent randomises deliberately over its actions in order to obtain more information about the environment (*exploration*) while, as the game proceeds, the agent becomes gradually more deterministic in its action choices so as to maximise its payoffs (*exploitation*). Typical mechanisms which implement this are $\epsilon$-greedy and Softmax/Boltzmann exploration (Sutton and Barto, 1998). Figure 1 demonstrates this in a SBG in which player $j$ has three reinforcement learning types. The payoffs for the types were such that the average overlap would eventually go to zero.

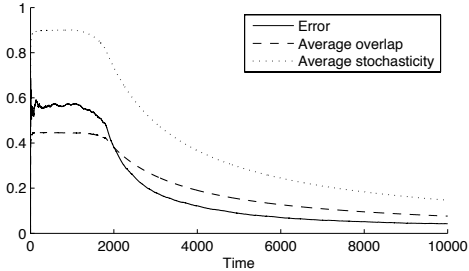

Figure 1: Example run in random SBG with 2 players, 10 actions, and 100 states. Player $j$ has three reinforcement learning types with $\epsilon$-greedy action selection (decreasing linearly from $\epsilon = 0.7$ at $t = 1000$, to $\epsilon = 0$ at $t = 2000$). The error at time $t$ is computed as $\sum_{\theta_j \in \Theta_j^+} |\text{Pr}_j(\theta_j|H^t) - \Upsilon(\theta_j)|$, where $\text{Pr}_j$ is the sum posterior.

Regarding the average overlap converging to zero, we believe that this is a property which should be guaranteed *by design*, for the following reason: If the hypothesised type space $\Theta_j^*$ is such that there is a constantly-high average overlap, then this means that the types in $\Theta_j^*$ are in effect very similar. However, types which are very similar are likely to produce very similar trajectories in the planning step of HBA (cf. $\hat{H}$ in (3)/(4)) and, therefore, constitute redundancy in both time and space. Thus, we believe it is advisable to use type spaces which have low average overlap.

*4.3. Correlated Posterior*

As noted earlier, an implicit assumption in (1) is that the type distribution $\Upsilon$ can be represented as a product of $n$ independent factors (one for each player), such that $\Upsilon(\theta) = \prod_j \Upsilon_j(\theta_j)$. Therefore, since the sum posterior is in the form of (1), it is in fact only guaranteed to learn independent type distributions. This is opposed to *correlated* type distributions, which cannot be represented as a product of $n$ independent factors. Correlated type distributions can be used to specify constraints on type combinations, such as "player $j$ can only have type $\theta_j$ if player $k$ has type $\theta_k$". The following example shows how the sum posterior may fail to converge to a correlated type distribution:

**Example 5.** Consider a SBG with 3 players. Player 1 is controlled by HBA using a sum posterior. Players 2 and 3 each have two types, $\Theta_2^+ = \Theta_3^+ = \{\theta_A, \theta_B\}$, which are defined as in Example 2. The type distribution $\Upsilon$ chooses types with probabilities $\Upsilon(\theta_A, \theta_B) = \Upsilon(\theta_B, \theta_A) = 0.5$ and $\Upsilon(\theta_A, \theta_A) = \Upsilon(\theta_B, \theta_B) = 0$. In other words, player 2 can never have the same type as player 3. From the perspective of HBA, each type (and hence action) is chosen with equal probability for both players. Thus, despite the fact that there is zero overlap and stochasticity, the sum posterior will eventually assign probability 0.25 to all constellations of types, which is incorrect. This means that HBA fails to recognise that the other players never choose the same action.

We now propose a posterior formulation which can learn any correlated type distribution:

**Definition 8.** The *correlated posterior* is defined as

$$\Pr(\theta^*_{-i}|H^t) = \eta\, P(\theta^*_{-i}) \sum_{\tau=0}^{t-1} \prod_{\theta^*_j \in \theta^*_{-i}} \pi_j(H^\tau, a^\tau_j, \theta^*_j) \tag{14}$$

where $P$ specifies prior beliefs over $\Theta^*_{-i}$ (analogous to $P_j$) and $\eta$ is a normaliser.

The correlated posterior is closely related to the sum posterior. In fact, it converges to the correct type distribution under the same conditions as the sum posterior:

**Theorem 3.** Let $\Gamma$ be a SBG with a *correlated* type distribution $\Upsilon$. If HBA uses the correlated posterior, then, for $t \to \infty$: If $\text{AO}_j(H^t) = 0$ and $\text{AS}_j(H^t) = 0$ for all players $j \neq i$, then $\Pr(\theta_{-i}|H^t) = \Upsilon(\theta_{-i})$ for all $\theta_{-i} \in \Theta^+_{-i}$.

*Proof.* The proof is analogous to the proof of Theorem 2. □

It is easy to see that the correlated posterior would learn the correct type distribution in Example 5. Note that, since it is guaranteed to learn any correlated type distribution, it is also guaranteed to learn any independent type distribution. Therefore, the correlated posterior would also learn the correct type distribution in Example 2. This means that the correlated posterior is *complete* in the sense that it covers the entire spectrum of pure/mixed and independent/correlated type distributions. However, this completeness comes at a higher computational complexity. While the sum posterior is in $O(n \max_j |\Theta^*_j|)$ time and space, the correlated posterior is in $O(\max_j |\Theta^*_j|^n)$ time and space. In practice, however, the time complexity can be reduced substantially by computing the probabilities $\pi_j(H^\tau, a^\tau_j, \theta^*_j)$ only once for each $j$ and $\theta^*_j \in \Theta^*_j$ (as in the sum posterior), and then reusing them in subsequent computations.

## 5. Practical Impact of Prior Beliefs

The previous section was concerned with the evolution of *posterior beliefs* as we observe more evidence. However, before we observe any evidence based on which to form our posterior beliefs, we will have to make an initial judgement as to the relative likelihood of types. This initial judgement is called the *prior belief*.

Given the lack of evidence, it may be tempting to use uniform prior beliefs in which all types have equal probability. Indeed, the fact that beliefs can change rapidly after only a few observations suggests that prior beliefs may have negligible effect. On the other hand, there is a substantial body of work in the game theory literature arguing the importance of prior beliefs (cf. Section 2). However, these works consider the impact of prior beliefs on equilibrium attainment when all players use the same type-based reasoning. In contrast, our interest is in the *practical* impact of prior beliefs, i.e. payoff maximisation, for a single agent using the type-based method.

In addition, there is the work of Bernardo (1979), Jaynes (1968), and others on "uninformed" priors. The purpose of such priors is to express a state of complete uncertainty, whilst possibly incorporating subjective prior information. (What this means and whether this is possible has been the subject of a long debate, e.g. De Finetti (2008).) However, this again differs from our interest in the impact of prior beliefs on payoff maximisation.

Thus, we are left with the following questions: Do prior beliefs have an impact on our ability to maximise payoffs in the long-term? If so, how? And, crucially, can we automatically compute prior beliefs so as to improve our long-term performance?

To find answers to these questions, we conducted a comprehensive empirical study which compared 10 methods to automatically compute prior beliefs from a given set of types. The results show that prior beliefs can indeed have a significant impact on the long-term performance, and that the depth of the planning horizon (i.e. how far we look into the future) plays a central role. Finally, and perhaps most intriguingly, we show that automatic methods can compute prior beliefs with consistent performance effects across a variety of scenarios. An implication of this is that prior beliefs could be eliminated as a manual parameter and instead be computed automatically.

The following subsections describe the experimental setup used in our study. The results are discussed in Section 5.7.

*5.1. Games*

We used a comprehensive set of benchmark games introduced by Rapoport and Guyer (1966), which consists of 78 repeated $2 \times 2$ matrix games. The games are *strictly ordinal*, meaning that each player ranks each of the four possible outcomes from 1 (least preferred) to 4 (most preferred), and no two outcomes have the same rank. Furthermore, the games are *distinct* in that no game can be obtained by transformation of any other game, which includes interchanging the rows, columns, and players (and any combination thereof) in the payoff matrix of the game.

The games can be grouped into 21 *no-conflict* games and 57 *conflict* games. In a no-conflict game, the two players have the same most preferred outcome, and so it is relatively easy to arrive at a solution that is best for both players. In a conflict game, the players disagree on the best outcome, hence they will have to find some form of a compromise.

We note that the games in this benchmark correspond to SBGs with *single* states. The simplicity of these games facilitates a thorough inspection of the interaction process and, thereby, explanation of observations. It also allows us to specify a *complete* benchmark set in the sense that it contains all games that satisfy the above description, which in turn allows us to draw more general conclusions. Finally, the fact that we use single-state games does not limit the inherent complexity of the interaction, since multi-state SBGs can always be emulated as single-state SBGs via an additional "nature" player (cf. Appendix A). Therefore, we expect that the principal observations we make will also hold in multi-state SBGs.

*5.2. Performance Criteria*

Each play of a game was partitioned into *time slices* which consist of an equal number of consecutive time steps. For each time slice, we measured the following performance criteria:

**Convergence** An agent converged in a time slice if its action probabilities in the time slice did not deviate by more than 0.05 from its initial action probabilities in the same time slice. Returns 1 (true) or 0 (false) for each agent.

**Average payoff** Average of payoffs an agent received in the time slice. Returns value in $[1, 4]$ for each agent.

**Welfare and fairness** Average sum and product, respectively, of the joint payoffs received in the time slice. Returns values in $[2, 8]$ and $[1, 16]$, respectively.

**Game solutions** Tests if the averaged action probabilities in the time slice formed an approximate stage-game Nash equilibrium, Pareto optimum, Welfare optimum, or Fairness optimum. Returns 1 (true) or 0 (false) for each game solution.

Precise formal definitions of these performance criteria can be found in (Albrecht and Ramamoorthy, 2012).

*5.3. Algorithm*

We used HBA to control player 1 and a fixed type in each play to control player 2, which was included in the set of hypothesised types $\Theta_2^*$ provided to HBA (discussed in detail in Section 5.6). Therefore, we used the product posterior formulation (cf. Section 4.1) to update HBA's beliefs. The planning step in HBA was implemented by expanding a finite tree of all future trajectories. Formally, HBA chooses an action $a_i$ which maximises the expected payoff $E_h^{a_i}(H^t)$, defined as

$$E_h^{a_i}(\hat{H}) = \sum_{\theta_j^* \in \Theta_j^*} \Pr{}_j(\theta_j^* \,|\, \hat{H}) \sum_{a_j \in A_j} \pi_j(\hat{H}, a_j, \theta_j^*) \, Q_{h-1}^{(a_i,a_j)}(\hat{H}) \qquad (15)$$

$$Q_h^{(a_i,a_j)}(\hat{H}) = u_i(a_i, a_j) + \begin{cases} 0 & \text{if } h = 0, \text{ else} \\ \max_{a_i'} E_h^{a_i'}\left(\langle \hat{H}, (a_i, a_j)\rangle\right) \end{cases} \qquad (16)$$

where $h$ specifies the depth of the planning horizon (i.e. HBA predicts the next $h$ actions of player $j$). Note that (15) and (16) correspond closely to (3) and (4), respectively. The difference is that (15)/(16) use $h$ to specify the planning depth while (3)/(4) use the discount factor $\gamma$. Hence, a "deeper" planning horizon $h$ translates into a greater discount factor $\gamma$. All results reported in this section hold for both variants.

*5.4. Types*

We used three different methods to automatically generate parameterised sets of types $\Theta_j^*$ for any given game. The generated types cover a broad spectrum of adaptive behaviours, including deterministic (CDT), randomised (CNN), and hybrid (LFT) policies. Algorithmic details and parameter settings can be found in Appendix B of (Albrecht, 2015).

**Leader-Follower-Trigger Agents (LFT)** Crandall (2014) described a method to generate sets of "leader" and "follower" agents which seek to play specific sequences of joint actions, called "target solutions". A leader agent plays its part of the target solution as long as the other player does. If the other player deviates, the leader agent punishes the player by playing a minimax strategy. The follower agent is similar except that it does not punish. Rather, if the other player deviates, the follower agent randomly resets its position within the target solution and continues play as usual. We augmented this set by a "trigger" agent which is similar to the leader and follower agents, except that it plays its maximin strategy indefinitely once the other player deviates.

**Co-Evolved Decision Trees (CDT)** We used genetic programming (Koza, 1992) to automatically breed sets of decision trees. A decision tree takes as input the past $n$ actions of the other player (in our case, $n = 3$) and deterministically returns an action to be played in response. The breeding process is co-evolutional, meaning that two pools of trees are bred concurrently (one for each player). In each evolution, a random selection of the trees for player 1 is evaluated against a random selection of the trees for player 2. The fitness criterion includes the payoffs generated by a tree as well as its dissimilarity to other trees in the same pool. This was done to encourage a more diverse breeding of trees, as otherwise the trees tend to become very similar or identical.

**Co-Evolved Neural Networks (CNN)** We used a string-based genetic algorithm (Holland, 1975) to breed sets of artificial neural networks. The process is basically the same as the one used for decision trees. However, the difference is that artificial neural networks can learn to play stochastic strategies while decision trees always play deterministic strategies. Our networks consist of one input layer with 4 nodes (one for each of the two previous actions of both players), a hidden layer with 5 nodes, and an output layer with 1 node. The node in the output layer specifies the probability of choosing action 1 (and, since we play $2 \times 2$ games, of action 2). All nodes use a sigmoidal threshold function and are fully connected to the nodes in the next layer.

### 5.5. Prior Beliefs

We specified a total of 10 different methods to automatically compute prior beliefs $P_j$ for a given set of types $\Theta_j^*$:

**Uniform prior** The uniform prior sets $P_j(\theta_j^*) = |\Theta_j^*|^{-1}$ for all $\theta_j^* \in \Theta_j^*$. This is the baseline prior against which the other priors are compared.

**Random prior** The random prior specifies $P_j(\theta_j^*) = .0001$ for a random half of the types in $\Theta_j^*$. The remaining probability mass is uniformly spread over the other half. The random prior is used to check if the performance differences of the various priors may be purely due to the fact that they concentrate the probability mass on fewer types.

**Value priors** Let $U_k^t(\theta_j^*)$ be the expected cumulative payoff to player $k$, from the start up until time $t$, if player $j$ (i.e. the other player) is of type $\theta_j^*$ and player $i$ (i.e. HBA) plays optimally against it. Each value prior is in the general form of $P_j(\theta_j^*) = \eta \psi(\theta_j^*)^b$, where $\eta$ is a normalisation constant and $b$ is a "booster" exponent used to magnify the differences between types $\theta_j^*$. Based on this general form, we define four different value priors:

- Utility prior: $\psi_U(\theta_j^*) = U_i^t(\theta_j^*)$
- Stackelberg prior: $\psi_S(\theta_j^*) = U_j^t(\theta_j^*)$
- Welfare prior: $\psi_W(\theta_j^*) = U_i^t(\theta_j^*) + U_j^t(\theta_j^*)$
- Fairness prior: $\psi_F(\theta_j^*) = U_i^t(\theta_j^*) * U_j^t(\theta_j^*)$

Our choice of value priors is motivated by the variety of metrics they cover. As a result, these priors can produce substantially different probabilities for the same set of types. In this study, we set $t = 5$ and $b = 10$.

**LP-priors** LP-priors are based on the idea that optimal priors can be formulated as the solution to a mathematical optimisation problem (in this case, a linear program). Each LP-prior generates a quadratic matrix $A$, where each element $A_{j,j'}$ contains the "loss" that HBA would incur if it planned its actions against the type $\theta_{j'}^*$ while the true type of player $j$ is $\theta_j^*$. Formally, let $U_k^t(\theta_j^*|\theta_{j'}^*)$ be like $U_k^t(\theta_j^*)$ except that HBA believes that player $j$ is of type $\theta_{j'}^*$ instead of $\theta_j^*$. We define four different LP-priors:

- LP-Utility: $A_{j,j'} = \psi_U(\theta_j^*) - U_i^t(\theta_j^*|\theta_{j'}^*)$
- LP-Stackelberg: $A_{j,j'} = \psi_S(\theta_j^*) - U_j^t(\theta_j^*|\theta_{j'}^*)$
- LP-Welfare: $A_{j,j'} = \psi_W(\theta_j^*) - [U_i^t(\theta_j^*|\theta_{j'}^*) + U_j^t(\theta_j^*|\theta_{j'}^*)]$
- LP-Fairness: $A_{j,j'} = \psi_F(\theta_j^*) - [U_i^t(\theta_j^*|\theta_{j'}^*) * U_j^t(\theta_j^*|\theta_{j'}^*)]$

The matrix $A$ can be fed into a linear program of the form $\min_c c^T x$ s.t. $[z, A]x \leq 0$, with $n = |\Theta_j^*|$, $c = (1, \{0\}^n)^T$, $z = (\{-1\}^n)^T$, to find a vector $x = (l, p_1, ..., p_n)$ in which $l$ is the minimised expected loss to HBA when using the probabilities $p_1, ..., p_n$ (one for each type) as the prior belief $P_j$. In order to avoid premature elimination of types, we furthermore require that $p_v > 0$ for all $1 \leq v \leq n$. As before, we set $t = 5$ and $b = 10$.

While this is a mathematically rigorous formulation, it is important to note that it is a simplification of how HBA really works. HBA incorporates its beliefs in every recursion of its planning procedure, whereas the LP formulation implicitly assumes that HBA uses its prior beliefs to randomly sample one of the types against which it then plans optimally. Nonetheless, this is often a reasonable approximation.

### 5.6. Experimental Procedure

We performed identical experiments for every type generation method described in Section 5.4. Each of the 78 games was played 10 times with different random seeds, and each play was repeated against three opponents (30 plays in total): **(RT)** A randomly generated type was used to control player 2 and the play lasted 100 rounds. **(FP)** A fictitious player (Brown, 1951) was used to control player 2 and the play lasted 10000 rounds. **(CFP)** A conditioned fictitious player (which learns action distributions conditioned on the previous joint action) was used to control player 2 and the play lasted 10000 rounds.

In each play, we randomly generated 9 unique types and provided them to HBA along with the true type of player 2, such that $|\Theta_2^*| = 10$. (That is, each play had a pure type distribution; cf. Section 3.1) Thus, the true type of player 2 was always included in the set of hypothesised types $\Theta_2^*$. To avoid "end-game" effects, the players were unaware the number of rounds. We included FP and CFP because they try to learn the behaviour of HBA. (While the generated types are adaptive, they do not create models of HBA's behaviour.) To facilitate the learning, we allowed for 10000 rounds. Finally, since FP and CFP will always choose dominating actions if they exist (in which case there is no interaction), we filtered out all games in the FP and CFP plays that had a dominating action for player 2 (leaving 15 no-conflict and 33 conflict games for the C/FP plays).

### 5.7. Results

We report three main observations:

**Observation 1.** *Prior beliefs can have a significant impact on the long-term performance of HBA.*

This was observed in all classes of types, against all classes of opponents, and in all classes of games used in this study. Figure 2 provides three representative examples from a range of scenarios. Many of the relative differences due to prior beliefs were statistically significant, based on paired two-sided t-tests with a 5% significance level.

Our data explain this as follows: Different prior beliefs may cause HBA to take different actions at the beginning of the game. These actions will shape the beliefs of the other player (i.e. how it models and adapts to HBA's actions) which in turn will affect HBA's next actions. Thus, if different prior beliefs lead to different initial actions, they may lead to different play trajectories with different payoffs.

Given that there is a time after which HBA will know the true type of player 2 (since it is provided to HBA), it may seem surprising that this process would lead to differences in the long-term. In fact, in our experiments, HBA often learned the true type after only 3 to 5 rounds, and in most cases in under 20 rounds. After that point, if the planning horizon of HBA is sufficiently deep, it will realise if its initial actions were sub-optimal and if it can manipulate the play trajectory to achieve higher payoffs in the long-term, thus diminishing the impact of prior beliefs.

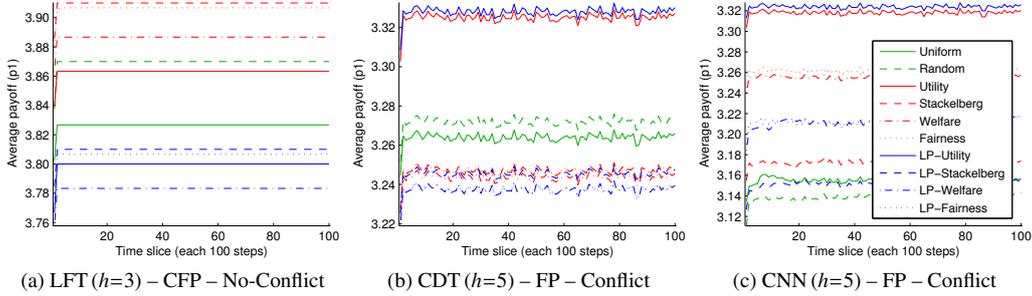

Figure 2: Prior beliefs can have significant impact on long-term performance. Plots show average payoffs of player 1 (HBA). X(*h*)–Y–Z format: HBA used X types and horizon *h*, player 2 was controlled by Y, results averaged over Z games.

However, deep planning horizons can be problematic in practice since the time complexity of HBA is exponential in the depth of the planning horizon. Therefore, the planning horizon constitutes a trade-off between decision quality and computational tractability. Interestingly, our data show that if we increase the depth, but stay below a sufficient depth ("sufficient" as described above), it may also *amplify* the impact of prior beliefs:

**Observation 2.** *Deeper planning horizons can diminish and amplify the impact of prior beliefs.*

Again, this was observed in all tested scenarios. Figures 3 and 4 show examples in which deeper planning horizons diminish and amplify the impact of prior beliefs, respectively.

How can deeper planning horizons amplify the impact of prior beliefs? Our data show that whether or not different prior beliefs cause HBA to take different initial actions depends not only on the prior beliefs and types, but also on the depth of the planning horizon. In some cases, differences between types (i.e. in their action choices) may be less visible in the near future and more visible in the distant future. In such cases, an HBA agent with a myopic planning horizon may choose the same (or similar) initial actions, despite different prior beliefs, because the differences in the types may not be visible within its planning horizon. On the other hand, an HBA agent with a deeper planning horizon may see the differences between the types and decide to choose different initial actions based on the prior beliefs.

We now turn to a comparison between the different prior beliefs. Here, our data reveal an intriguing property:

**Observation 3.** *Automatic methods can compute prior beliefs with consistent performance effects.*

Figure 5 shows that the prior beliefs had consistent performance effects across a wide variety of scenarios. For example, the Utility prior produced consistently higher payoffs for player 1 (i.e. HBA) while the Stackelberg prior produced consistently higher payoffs for player 2 as well as higher welfare and fairness. The Welfare and Fairness priors were similar to the Stackelberg prior, but not quite as consistent. Similar results were observed for the LP variants of the priors, despite the fact that the LP formulation is a simplification of how HBA works (cf. Section 5.5).

We note that none of the prior beliefs, including the Uniform prior, produced high rates for the game solutions (i.e. Nash equilibrium, Pareto optimality, etc.). This is because we measured *stage-game* solutions, which have no notion of time. These can be hard to attain in repeated games, especially if the other player does not actively seek a specific solution, as was often the case in our study.

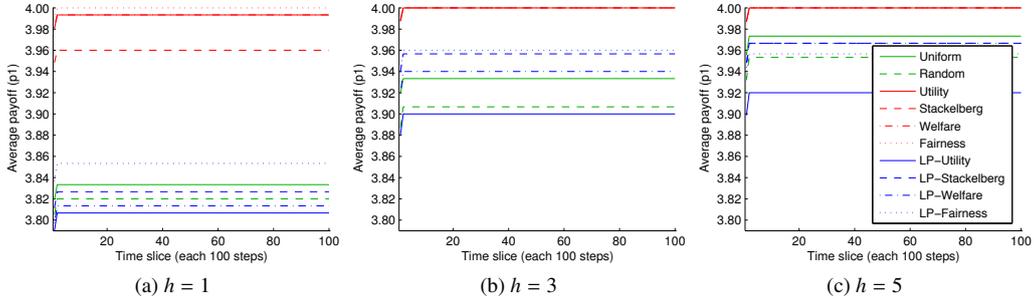

Figure 3: Deeper planning horizons can diminish impact of prior beliefs. Results shown for HBA with LFT types, player 2 controlled by FP, averaged over no-conflict games. *h* is depth of planning horizon (predicting *h* next actions of player 2).

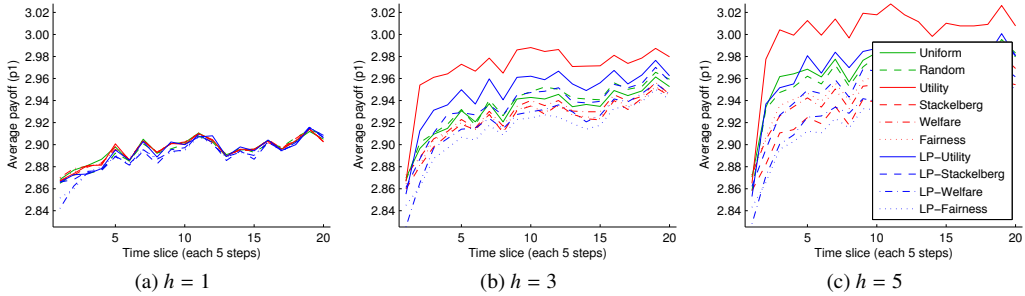

Figure 4: Deeper planning horizons can amplify impact of prior beliefs. Results shown for HBA with CNN types, player 2 controlled by RT, averaged over conflict games. *h* is depth of planning horizon (predicting *h* next actions of player 2).

Observation 3 is intriguing because it indicates that prior beliefs could be eliminated as a manual parameter and instead be computed automatically, using methods such as the ones specified in Section 5.5. The fact that our methods produced consistent results means that prior beliefs can be constructed to optimise specific performance criteria. Note that this result is particularly interesting because the prior beliefs have no influence, whatsoever, on the true type of player 2.

This observation is further supported by the fact that the Random prior did not produce consistently different values (for any criterion) from the Uniform prior. This means that the differences in the prior beliefs are not merely due to the fact that they concentrate the probability mass on fewer types, but rather that the prior beliefs reflect the intrinsic metrics based on which they are computed (e.g. player 1 payoffs for Utility prior, player 2 payoffs for Stackelberg prior).

How is this phenomenon explained? We believe this may be an interesting analogy to the "optimism in uncertainty" principle (e.g. Brafman and Tennenholtz, 2003). The optimism lies in the fact that HBA commits to a specific class of types – those with high prior belief – while, in truth and without further evidence, there is no reason to believe that any one type is a priori more likely than others.

Each class of types is characterised by the intrinsic metric of the prior belief. For instance, the Utility prior assigns high probability to those types which would yield high payoffs to HBA if it played optimally against the types. By committing to such a characterisation, HBA can effectively utilise Observation 1 by choosing initial actions so as to shape the interaction to maximise the intrinsic metric. If the true type of player 2 is indeed in this class of types, then the interaction will proceed as planned by HBA and the intrinsic metric will be optimised. However, if the true type is not in this class, then HBA will quickly learn the correct type and adjust its play accordingly, albeit without necessarily maximising the intrinsic metric.

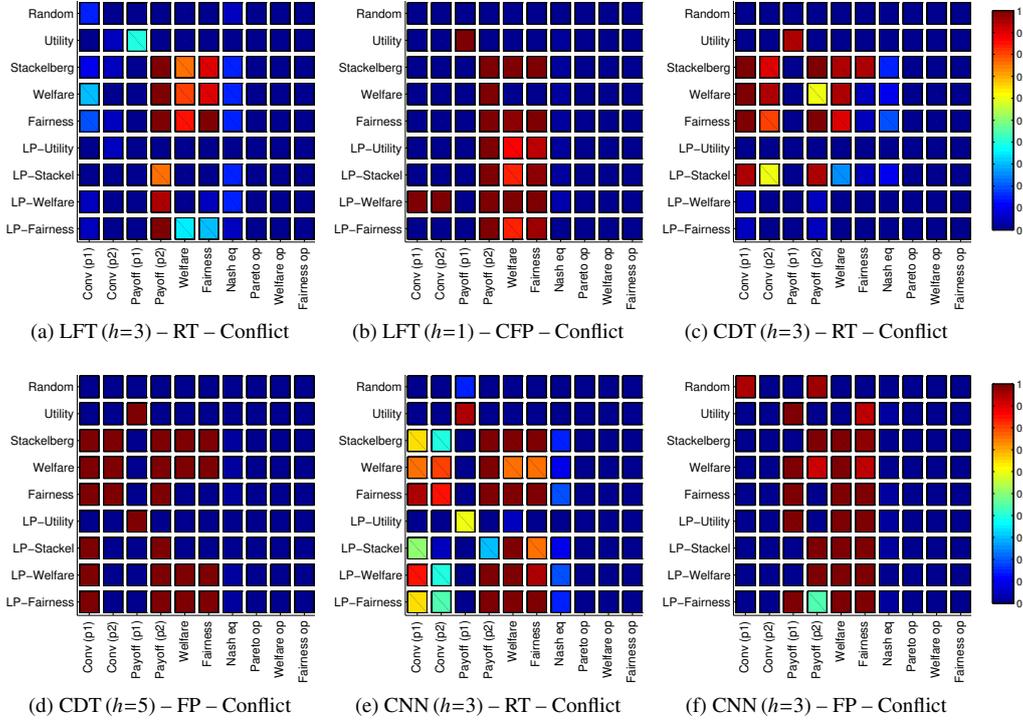

Figure 5: Automatic prior beliefs have consistent performance effects. Rows show prior beliefs, columns show performance criteria. Each element $(r,c)$ in the matrix corresponds to the percentage of time slices in which the prior belief $r$ produced significantly higher values for criterion $c$ than the Uniform prior, averaged over all plays in all tested games. All significance statements are based on paired right-sided t-tests with 5% significance level. See Figure 2 for X($h$)–Y–Z format.

This is in contrast to the Uniform and Random priors, which have no intrinsic metric. Under these priors, HBA will plan its actions with respect to types which are not characterised by a common theme (i.e., all types under the Uniform prior, and a random half under the Random prior). Therefore, HBA cannot effectively utilise Observation 1.

## 6. Optimal Type Spaces

A potential concern in the type-based method is the fact that the hypothesised types may be *incorrect*. This can range from slight deviations in predicted action probabilities, to predicting entirely different actions from what was observed. The following example illustrates this:

**Example 6.** Consider a SBG with two players and actions L and R. Player 1 is controlled by HBA while player 2 has a single type, $\theta_{LR}$, which chooses L,R,L,R, etc. HBA is provided with hypothesised types $\Theta_j^* = \left\{\theta_R^*, \theta_{LRR}^*\right\}$, where $\theta_R^*$ always chooses R while $\theta_{LRR}^*$ chooses L,R,R,L,R,R etc. Both hypothesised types are incorrect in the sense that they predict player 2's actions in only $\approx 50\%$ of the game.

Such inaccuracies may have a significant impact on our choice of actions: if the hypothesised types are incorrect, then our predictions of future interactions may be incorrect, which in turn may lead to suboptimal action choices. Therefore, an important question is what relation the hypothesised types must have to the true types in order for HBA to be able to complete its task? In particular, what does it mean for the hypothesised types to be *optimal*?

Given the complexity of behaviours agents may exhibit, this is an extremely difficult question. In addition, it is not generally sufficient to consider types alone, since actions are planned with respect to both types *and* beliefs over types. Rather, we have to consider a stochastic process in which our actions depend on the correctness of types as well as the evolution of our beliefs.

In this spirit, we describe a formal methodology whereby we compare two interactive processes: one in which the true types are known, and one in which this knowledge is approximated through beliefs over hypothesised types. Based on these processes, we use a probabilistic temporal logic to define a hierarchy of desirable termination guarantees, and analyse the theoretical conditions under which they are met. The main result of this analysis is a novel characterisation of optimality which is based on the concept of probabilistic bisimulation (Larsen and Skou, 1991). In addition to concisely defining what constitutes optimality of hypothesised types, this allows the user to apply efficient model checking algorithms to verify optimality in practice.

### 6.1. Task Completion

We are interested in *task completion*, which we formally capture by the following assumption:

**Assumption 2.** Let player $i$ be controlled by HBA. Then $u_i(s, a) = 1$ iff. $s \in \bar{S}$, else 0.

Assumption 2 specifies that we are only interested in reaching a terminal state, since this is the only way to obtain a none-zero payoff. In our analysis, we consider discount factors $\gamma$ (cf. Algorithm 1) with $\gamma = 1$ and $\gamma < 1$. While all our results hold for both cases, there is an important distinction: If $\gamma = 1$, then the expected payoffs (3) correspond to the actual probability that the following state can lead to (or is) a terminal state (we call this the *success rate*), whereas this is not necessarily the case if $\gamma < 1$. This is since $\gamma < 1$ tends to prefer shorter paths, which means that actions with lower success rates may be preferred if they lead to faster termination. Therefore, if $\gamma = 1$ then HBA is solely interested in termination, and if $\gamma < 1$ then it is interested in *fast* termination, where lower $\gamma$ prefers faster termination.

### 6.2. Methodology of Analysis

Given a SBG $\Gamma$, we define the *ideal process*, $X$, as the process induced by $\Gamma$ in which player $i$ is controlled by HBA and in which HBA always knows the current and all future types of all players. Then, given a posterior formulation Pr and hypothesised type spaces $\Theta_j^*$ for all $j \neq i$, we define the *user process*, $Y$, as the process induced by $\Gamma$ in which player $i$ is controlled by HBA (same as in $X$) and in which HBA uses Pr and $\Theta_j^*$ in the usual way. Thus, the only difference between $X$ and $Y$ is that $X$ can always predict the player types whereas $Y$ approximates this knowledge through Pr and $\Theta_j^*$. We write $E_{s^t}^{a_i}(H^t|C)$ to denote the expected payoff (as defined by (3)) of action $a_i$ in state $s^t$ after history $H^t$, in process $C \in \{X, Y\}$.

The idea is that $X$ constitutes the ideal solution in the sense that $E_{s^t}^{a_i}(H^t|X)$ corresponds to the *actual* expected payoff, which means that HBA chooses the truly best-possible actions in $X$. This is opposed to $E_{s^t}^{a_i}(H^t|Y)$, which is merely the *estimated* expected payoff based on Pr and $\Theta_j^*$, so that HBA may choose suboptimal actions in $Y$. The methodology of our analysis is to specify what relation $Y$ must have to $X$ to satisfy certain guarantees for termination.

We specify such guarantees in PCTL (Hansson and Jonsson, 1994), a probabilistic modal logic which also allows for the specification of time constraints. PCTL expressions are interpreted over infinite histories in labelled transition systems with atomic propositions (i.e. Kripke structures). In order to interpret PCTL expressions over $X$ and $Y$, we make the following modifications without loss of generality: Firstly, any terminal state $\bar{s} \in \bar{S}$ is an *absorbing* state, meaning that if a process is in $\bar{s}$, then the next state will be $\bar{s}$ with probability 1 and all players receive a zero payoff. Secondly, we introduce the atomic proposition term and label each terminal state with it, so that term is true in $s$ if and only if $s \in \bar{S}$.

We will use the following two PCTL expressions:

$$F_{>p}^{\leq t}\texttt{term},\ F_{>p}^{<\infty}\texttt{term} \tag{17}$$

where $t \in \mathbb{N}$, $p \in [0, 1]$, and $> \in \{>, \geq\}$.

$F_{>p}^{\leq t}\texttt{term}$ specifies that, given a state $s$, with a probability of $>p$ a state $s'$ will be reached from $s$ within $t$ time steps such that $s'$ satisfies term. The semantics of $F_{>p}^{<\infty}\texttt{term}$ is similar except that $s'$ will be reached in arbitrary but finite time. We write $s \models_C \phi$ to say that a state $s$ satisfies the PCTL expression $\phi$ in process $C \in \{X, Y\}$.

*6.3. Critical Type Spaces*

In our analysis, we will sometimes assume that the hypothesised type spaces $\Theta_j^*$ are *uncritical*:

**Definition 9.** The hypothesised type spaces $\Theta_j^*$ are *critical* if there is a set $S^c \subseteq S \setminus \bar{S}$ which satisfies all of the following:

1. For each $H^t \in \mathbb{H}$ with $s^t \in S^c$, there is $a_i \in A_i$ such that $E_{s^t}^{a_i}(H^t|Y) > 0$ and $E_{s^t}^{a_i}(H^t|X) > 0$.
2. There is a positive probability that $Y$ may eventually get into a state $s^c \in S^c$ from $s^0$.
3. If $Y$ is in a state in $S^c$, then with probability 1 it will always be in a state in $S^c$.

We say $\Theta_j^*$ are *uncritical* if they are not critical.

Intuitively, critical type spaces have the potential to lead HBA into a state space in which it *believes* it chooses the right actions to complete the task, while other actions are *actually* required to complete the task. The only effect that its actions have is to induce an infinite cycle, due to a critical inconsistency between the hypothesised and true type spaces. The following example demonstrates this:

**Example 7.** Recall Example 6 and let the task be to choose the same action as player $j$. Then, $\Theta_j^*$ is uncritical because HBA will always complete the task at $t = 1$, regardless of its posterior beliefs and despite the fact that $\Theta_j^*$ is inaccurate. Now, assume that $\Theta_j^* = \{\theta_{RL}^*\}$ where $\theta_{RL}^*$ chooses actions R,L,R,L etc. Then, $\Theta_j^*$ is critical since HBA will always choose the opposite action of player $j$, thinking that it would complete the task, when a different action would actually complete it.

A practical way to ensure that the type spaces $\Theta_j^*$ are (eventually) uncritical is to include methods for opponent modelling in each $\Theta_j^*$. If the opponent models are guaranteed to learn the correct behaviours, then the type spaces $\Theta_j^*$ are guaranteed to become uncritical. In Example 7, any standard modelling method would eventually learn that the true strategy of player $j$ is $\theta_{LR}$. As the model becomes more accurate, the posterior beliefs gradually shift towards it and eventually allow HBA to take the right action.

*6.4. Termination Guarantees*

Our first termination guarantee states that if $X$ has a positive probability of solving the task, then so does $Y$:

**Property 1.** $s^0 \models_X F_{>0}^{<\infty}\texttt{term} \Rightarrow s^0 \models_Y F_{>0}^{<\infty}\texttt{term}$

We can show that Property 1 holds if the hypothesised type spaces $\Theta_j^*$ are uncritical and if $Y$ only chooses actions for player $i$ with positive expected payoff in $X$.

Let $\mathbb{A}(H^t|C)$ denote the set of actions that process $C$ may choose from in state $s^t$ after history $H^t$, i.e. $\mathbb{A}(H^t|C) = \arg\max_{a_i} E_{s^t}^{a_i}(H^t|C)$ (cf. step 3 in Algorithm 1).

**Theorem 4.** Property 1 holds if $\Theta_j^*$ are uncritical and

$$\forall H^t \in \mathbb{H} \; \forall a_i \in \mathbb{A}(H^t|Y) : E_{s^t}^{a_i}(H^t|X) > 0 \tag{18}$$

*Proof.* Assume $s^0 \models_X F_{>0}^{<\infty}\texttt{term}$. Then, we know that $X$ chooses actions $a_i$ which *may* lead into a state $s'$ such that $s' \models_X F_{>0}^{<\infty}\texttt{term}$, and the same holds for all such states $s'$. Now, given (18) it is tempting to infer the same result for $Y$, since $Y$ only chooses actions $a_i$ which have positive expected payoff in $X$ and, therefore, could truly lead into a terminal state. However, (18) alone is not sufficient to infer $s' \models_Y F_{>0}^{<\infty}\texttt{term}$ because of the special case in which $Y$ chooses actions $a_i$ such that $E_{s^t}^{a_i}(H^t|X) > 0$ but without ever reaching a terminal state. This is why we require that the hypothesised type spaces $\Theta_j^*$ are uncritical, which prevents this special case. Thus, we can infer that $s' \models_Y F_{>0}^{<\infty}\texttt{term}$, and, hence, Property 1 holds. $\square$

The second guarantee states that if $X$ always completes the task, then so does $Y$:

**Property 2.** $s^0 \models_X F_{\geq 1}^{<\infty}\texttt{term} \Rightarrow s^0 \models_Y F_{\geq 1}^{<\infty}\texttt{term}$

We can show that Property 2 holds if the type spaces $\Theta_j^*$ are uncritical and if $Y$ only chooses actions for player $i$ which lead to states into which $X$ may get as well.

Let $\mu(H^t, s|C)$ be the probability that process $C$ transitions into state $s$ from state $s^t$ after history $H^t$, i.e.

$$\mu(H^t, s|C) = \frac{1}{|\mathbb{A}|} \sum_{a_i \in \mathbb{A}} \sum_{a_{-i} \in A_{-i}} T(s^t, \langle a_i, a_{-i} \rangle, s) \prod_{j \neq i} \pi_j(H^t, a_j, \theta_j^t) \tag{19}$$

with $\mathbb{A} \equiv \mathbb{A}(H^t|C)$, and let $\mu(H^t, S'|C) = \sum_{s \in S'} \mu(H^t, s|C)$ for $S' \subset S$.

**Theorem 5.** Property 2 holds if $\Theta_j^*$ are uncritical and

$$\forall H^t \in \mathbb{H} \; \forall s \in S : \mu(H^t, s|Y) > 0 \Rightarrow \mu(H^t, s|X) > 0 \tag{20}$$

*Proof.* The fact that $s^0 \models_X F_{\geq 1}^{<\infty}\texttt{term}$ means that, throughout the process, $X$ only transitions into states $s$ with $s \models_X F_{\geq 1}^{<\infty}\texttt{term}$. As before, it is tempting to infer the same result for $Y$ based on (20), since it only transitions into states which have maximum success rate in $X$. However, (20) alone is not sufficient since $Y$ may choose actions such that (20) holds true but $Y$ will never reach a terminal state. Nevertheless, since the hypothesised type spaces $\Theta_j^*$ are uncritical, we know that this special case will not occur, and, thus, Property 2 holds. □

We note that, in both Properties 1 and 2, the reverse direction holds true regardless of Theorems 4 and 5. Furthermore, we can combine the requirements of Theorems 4 and 5 to ensure that both properties hold.

The third guarantee subsumes the previous guarantees by stating that $X$ and $Y$ have the same minimum probability of solving the task:

**Property 3.** $s^0 \models_X F_{\geq p}^{<\infty}\texttt{term} \Rightarrow s^0 \models_Y F_{\geq p}^{<\infty}\texttt{term}$

We can show that Property 3 holds if the hypothesised type spaces $\Theta_j^*$ are uncritical and if $Y$ only chooses actions for player $i$ which $X$ might have chosen as well.

Let $R(a_i, H^t|C)$ be the *success rate* of action $a_i$, formally $R(a_i, H^t|C) = E_{s^t}^{a_i}(H^t|C)$ with $\gamma = 1$ (so that it corresponds to the actual *probability* with which $a_i$ may lead to termination in the future). Define $X_{\min}$ and $X_{\max}$ to be the processes which for each $H^t$ choose actions $a_i \in \mathbb{A}(H^t|X)$ with, respectively, minimal and maximal success rate $R(a_i, H^t|X)$.

**Theorem 6.** If $\Theta_j^*$ are uncritical and

$$\forall H^t \in \mathbb{H} : \mathbb{A}(H^t|Y) \subseteq \mathbb{A}(H^t|X) \tag{21}$$

then
(i) for $\gamma = 1$: Proposition 3 holds in both directions
(ii) for $\gamma < 1$: $s^0 \models_X F_{\geq p}^{<\infty}\texttt{term} \Rightarrow s^0 \models_Y F_{\geq p'}^{<\infty}\texttt{term}$
with $p_{\min} \leq q \leq p_{\max}$ for $q \in \{p, p'\}$, where $p_{\min}$ and $p_{\max}$ are the highest probabilities such that $s^0 \models_{X_{\min}} F_{\geq p_{\min}}^{<\infty}\texttt{term}$ and $s^0 \models_{X_{\max}} F_{\geq p_{\max}}^{<\infty}\texttt{term}$.

*Proof.* (i): Since $\gamma = 1$, all actions $a_i \in \mathbb{A}(H^t|X)$ have the same success rate for a given $H^t$, and given (21) we know that $Y$'s actions always have the same success rate as $X$'s actions. Provided that the type spaces $\Theta_j^*$ are uncritical, we can conclude that Property 3 must hold, and for the same reasons the reverse direction must hold as well.

(ii): Since $\gamma < 1$, the actions $a_i \in \mathbb{A}(H^t|X)$ may have different success rates. The lowest and highest chances that $X$ completes the task are precisely modelled by $X_{\min}$ and $X_{\max}$, and given (21) and the fact that $\Theta_j^*$ are uncritical, the same holds for $Y$. Therefore, we can infer the common bound $p_{\min} \leq \{p, p'\} \leq p_{\max}$ as defined in Theorem 6. □

Properties 1 to 3 are *indefinite* in the sense that they make no restrictions on time requirements. Our fourth and final guarantee subsumes all previous guarantees and states that if there is a probability $p$ such that $X$ terminates *within $t$ time steps*, then so does $Y$ for the same $p$ and $t$:

**Property 4.** $s^0 \models_X F_{\geq p}^{\leq t}\texttt{term} \Rightarrow s^0 \models_Y F_{\geq p}^{\leq t}\texttt{term}$

We believe that Property 4 is an adequate criterion of *optimality* for hypothesised type spaces $\Theta_j^*$ since, if it holds, $\Theta_j^*$ must approximate the true type spaces $\Theta_j^+$ in a way which allows HBA to plan (almost) as accurately — in terms of solving the task — as the "ideal" HBA in $X$ which always knows the true types.

What relation must $Y$ have to $X$ in order to satisfy Property 4? The fact that $Y$ and $X$ are processes over state transition systems means that we can draw on methods from the model checking literature to answer this question. Specifically, we will use the concept of *probabilistic bisimulation* (Larsen and Skou, 1991), which we here define within the context of our work:

**Definition 10.** A *probabilistic bisimulation* between $X$ and $Y$, denoted $X \sim Y$, is an equivalence relation $B \subseteq S \times S$ such that
    (i) $(s^0, s^0) \in B$
    (ii) $s_X \models_X \mathtt{term} \Leftrightarrow s_Y \models_Y \mathtt{term}$ for all $(s_X, s_Y) \in B$
    (iii) $\mu(H_X^t, \hat{S}|X) = \mu(H_Y^t, \hat{S}|Y)$ for any histories $H_X^t, H_Y^t$ with $(s_X^t, s_Y^t) \in B$ and all equivalence classes $\hat{S}$ under $B$.

Intuitively, a probabilistic bisimulation states that $X$ and $Y$ do (on average) match each other's transitions. Our definition of probabilistic bisimulation is most general in that it does not require that transitions are matched by the same action or that related states satisfy the same atomic propositions other than termination. However, we do note that other definitions exist which make such additional requirements, and our results hold for each of these refinements.

The main contribution in this section is to show that the optimality criterion expressed by Property 4 holds in *both directions* if there exists a probabilistic bisimulation between $X$ and $Y$. Thus, we offer an alternative formal characterisation of optimality for the hypothesised type spaces $\Theta_j^*$:

**Theorem 7.** Property 4 holds in both directions if there exists a probabilistic bisimulation $X \sim Y$.

*Proof.* First of all, we note that, strictly speaking, the standard definitions of bisimulation (e.g. Baier, 1996; Larsen and Skou, 1991) assume the Markov property, which means that the next state of a process depends only on its current state. In contrast, we consider the more general case in which the next state may depend on the history $H^t$ of previous states and joint actions (since the player strategies $\pi_j$ depend on $H^t$). However, one can always enforce the Markov property *by design*, i.e. by augmenting the state space $S$ to account for the relevant factors of the past. In fact, we could postulate that the histories as a whole constitute the states of the system, i.e. $S = \mathbb{H}$. Therefore, to simplify the exposition, we assume the Markov property and we write $\mu(s, \hat{S}|C)$ to denote the cumulative probability that $C$ transitions from state $s$ into any state in $\hat{S}$.

Given the Markov property, the fact that $B$ is an equivalence relation, and $\mu(s_X, \hat{S}|X) = \mu(s_Y, \hat{S}|Y)$ for $(s_X, s_Y) \in B$, we can represent the dynamics of $X$ and $Y$ in a common graph, such as the following one:

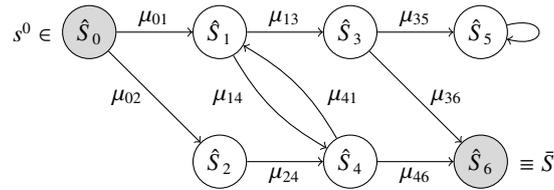

The nodes correspond to the equivalence classes under $B$. A directed edge from $\hat{S}_a$ to $\hat{S}_b$ specifies that there is a positive probability $\mu_{ab} = \mu(s_X, \hat{S}_b|X) = \mu(s_Y, \hat{S}_b|Y)$ that $X$ and $Y$ transition from states $s_X, s_Y \in \hat{S}_a$ to states $s_X', s_Y' \in \hat{S}_b$, respectively. Note that $s_X, s_Y$ and $s_X', s_Y'$ need not be equal but merely equivalent, i.e. $(s_X, s_Y) \in B$ and $(s_X', s_Y') \in B$. There is one node ($\hat{S}_0$) that contains the initial state $s^0$ and one node ($\hat{S}_6$) that contains all terminal states $\bar{S}$ and no other states. This is because once $X$ and $Y$ reach a terminal state they will always stay in it (i.e. $\mu(s, \bar{S}|X) = \mu(s, \bar{S}|Y) = 1$ for $s \in \bar{S}$) and since they are the only states that satisfy $\mathtt{term}$. Thus, the graph starts in $\hat{S}_0$ and terminates (if at all) in $\hat{S}_6$.

Since the graph represents the dynamics of both *X* and *Y*, it is easy to see that Property 4 must hold in both directions. In particular, the probabilities that *X* and *Y* are in node $\hat{S}$ at time *t* are identical. One simply needs to add the probabilities of all directed paths of length *t* which end in $\hat{S}$ (provided that such paths exist), where the probability of a path is the product of the $\mu_{ab}$ along the path. Therefore, *X* and *Y* terminate with equal probability, and on average within the same number of time steps. □

Some remarks to clarify the usefulness of this result: First of all, in contrast to Theorems 4 to 6, Theorem 7 does not explicitly require $\Theta_j^*$ to be uncritical. In fact, this is implicit in the definition of probabilistic bisimulation. Moreover, while the other theorems relate *Y* and *X* for identical histories $H^t$, Theorem 7 relates *Y* and *X* for *related* histories $H_Y^t$ and $H_X^t$, making it more generally applicable. Finally, Theorem 7 has an important practical implication: it tells us that we can use efficient methods for model checking (e.g. Baier, 1996; Larsen and Skou, 1991) to verify optimality of $\Theta_j^*$. In fact, it can be shown that for Property 4 to hold (albeit not in the other direction) it suffices that *Y* be a *probabilistic simulation* (Baier, 1996) of *X*, which is a coarser preorder than probabilistic bisimulation. However, algorithms for checking probabilistic simulation (e.g. Baier, 1996) are computationally much more expensive (and fewer) than those for probabilistic bisimulation, hence their practical use is currently limited.

## 7. Behavioural Hypothesis Testing

In the previous section, we considered the possibility of incorrect hypothesised types and analysed the conditions under which HBA is nevertheless able to complete its task. While the analysis is rigorous and complete, it is performed *before* any interaction and with respect to the true types of other agents. How can we decide *during* the interaction and with no knowledge of the true types whether our hypothesised types are correct?

There are several ways in which an answer to this question could be used. For example, if we persistently reject our hypothesised types, we may hypothesise an alternative set of types or resort to some default plan of action, such as a "maximin" strategy. Unfortunately, posterior beliefs do not provide an answer to this question because they quantify the *relative* likelihood of types (relative to a set of alternative types), but they are no measure of truth. That is, even if our beliefs point to one type, this does not tell us that the observed agent is indeed of that type. Instead, it only tells us that all other types have been discarded after the current interaction history.

To illustrate the source of difficulty, consider an interaction process between two agents which can choose from three actions. The table below shows the first 5 time steps of the interaction. The columns show, respectively, the current time *t* of the interaction, the actions chosen by the agents at time *t*, and agent 1's hypothesised probabilities with which agent 2 will choose its actions at time *t*, based on the prior interaction history.

| *t* | $(a_1^t, a_2^t)$ | $\theta_2^*$ |
|---|---|---|
| 1 | (1, 2) | $\langle .3, .1, .6 \rangle$ |
| 2 | (3, 1) | $\langle .2, .3, .5 \rangle$ |
| 3 | (2, 3) | $\langle .7, .1, .2 \rangle$ |
| 4 | (2, 3) | $\langle .0, .4, .6 \rangle$ |
| 5 | (1, 2) | $\langle .4, .2, .4 \rangle$ |

Assuming the process continues in this fashion, and without any restrictions on the behaviour of agent 2, how should agent 1 decide whether or not to reject its hypothesis about the behaviour of agent 2? Note that agent 1 cannot outright reject its hypothesis because all observed actions of agent 2 were supported by agent 1's hypothesis (i.e. had positive probability).

There exists a large body of literature on what is often referred to as *model criticism* (e.g. Bayarri and Berger, 2000; Meng, 1994; Rubin, 1984; Box, 1980). Model criticism attempts to answer the analogous question of whether a given data set could have been generated by a given model. However, in contrast to our work, model criticism usually assumes that the data are independent and identically distributed, which is not the case in the interactive setting we consider. A related problem, sometimes referred to as *identity testing*, is to test if a given sequence of data was generated by some given stochastic process (Ryabko and Ryabko, 2008; Basawa and Scott, 1977). Instead of independent and identical distributions, this line of work assumes other properties such as stationarity and ergodicity. Unfortunately, these assumptions are also unlikely in interaction processes, and the proposed solutions are very costly.

A perhaps more natural way to address this question is to compute some kind of *score* from the information given in the above table, and to compare this score with some manually chosen rejecting threshold. A prominent example of such a score is the empirical frequency distribution (e.g. Conitzer and Sandholm, 2007; Foster and Young, 2003). However, while the simplicity of this method is appealing, there are two significant problems: (a) it is far from trivial to devise a scoring scheme that reliably quantifies "correctness" of hypotheses (for instance, an empirical frequency distribution taken over all past actions would be insufficient in the above example since the hypothesised action distributions are changing), and (b) it is unclear how one should choose the threshold parameter for any given scoring scheme.

In this section, we show how a particular form of model criticism, namely frequentist hypothesis testing, can be combined with the concept of scores to decide whether to reject a behavioural hypothesis. Our proposed algorithm addresses (a) by allowing for multiple scoring criteria in the construction of the test statistic, with the intent of obtaining an overall more reliable scoring scheme. The distribution of the test statistic is learned during the interaction process, and we show that the learning is asymptotically correct. Analogous to standard frequentist testing, the hypothesis is rejected at a given point in time if the resulting *p*-value is below some "significance level". This eliminates (b) by providing a uniform semantics for rejection that is invariant to the employed scoring scheme. We present results from a comprehensive set of experiments, demonstrating that the algorithm achieves high accuracy and scalability at low computational costs.

Of course, there is a long-standing debate on the role of statistical hypothesis tests and quantities such as *p*-values (e.g. Gelman and Shalizi, 2013; Berger and Sellke, 1987; Cox, 1977). The usual consensus is that *p*-values should be combined with other forms of evidence to reach a final conclusion (Fisher, 1935), and this is the view we adopt as well. In this sense, our method may be used as part of a larger machinery to decide the truth of a hypothesis.

### 7.1. Individual Hypotheses and Beliefs

As noted in Section 6, it does not generally suffice to consider the correctness of individual types, since we plan our actions with respect to both types *and* our beliefs regarding the relative likelihood of types (cf. (3)). In this regard, we note that any combination of beliefs Pr and types $\Theta_j^*$ can be described as a single type $\hat{\theta}_j^*$ of the form

$$\pi_j(H^t, a_j, \hat{\theta}_j^*) = \sum_{\theta_j^* \in \Theta_j^*} \Pr(\theta_j^* | H^t) \pi_j(H^t, a_j, \theta_j^*). \tag{22}$$

This combination is equivalent to sampling a single type $\theta_j^* \in \Theta_j^*$ using probabilities $\Pr(\theta_j^*|H^t)$, and then using $\theta_j^*$ to choose actions $a_j \in A_j$ via $\pi_j(H^t, a_j, \theta_j^*)$ (Kuhn, 1953). Analogously, we may combine the true types $\Theta_j^+ \subset \Theta_j$ of player $j$, using the type distribution $\Upsilon$, into a single type $\hat{\theta}_j^+$ such that

$$\pi_j(H^t, a_j, \hat{\theta}_j^+) = \sum_{\theta_j^+ \in \Theta_j^+} \Upsilon(t, \theta_j^+) \pi_j(H^t, a_j, \theta_j^+). \tag{23}$$

Therefore, to simplify the notation in this section, we will generally assume a single hypothesised type $\theta_j^* \in \Theta_j$ and a single true type $\theta_j^+ \in \Theta_j$. Note that this means that our method can be applied to the combination of beliefs and hypothesised types, as well as to individual types in $\Theta_j^*$. Furthermore, we will write $\pi_j(H^t, \theta_j)$ to denote the probability distribution over actions $A_j$ (rather than probabilities of individual actions).

### 7.2. A Method for Behavioural Hypothesis Testing

Let $i$ denote our agent and let $j$ denote another agent. Moreover, let $\theta_j^* \in \Theta_j$ denote our hypothesis for $j$'s behaviour and let $\theta_j^+ \in \Theta_j$ denote $j$'s true behaviour. The central question we ask is if $\theta_j^* = \theta_j^+$?

Unfortunately, since we do not know $\theta_j^+$, we cannot directly answer this question. However, at each time $t$, we know $j$'s past actions $\mathbf{a}_j^t = (a_j^0, ..., a_j^{t-1})$ which were generated by $\theta_j^+$. If we use $\theta_j^*$ to generate a vector $\hat{\mathbf{a}}_j^t = (\hat{a}_j^0, ..., \hat{a}_j^{t-1})$, where $\hat{a}_j^\tau$ is sampled using $\pi_j(H^\tau, \theta_j^*)$, we can formulate the related two-sample problem of whether $\mathbf{a}_j^t$ and $\hat{\mathbf{a}}_j^t$ were generated from the same behaviour, namely $\theta_j^*$.

In this section, we propose a general and efficient algorithm to decide this problem. At its core, the algorithm computes a frequentist $p$-value

$$p = P\left(|T(\tilde{\mathbf{a}}_j^t, \hat{\mathbf{a}}_j^t)| \geq |T(\mathbf{a}_j^t, \hat{\mathbf{a}}_j^t)|\right) \tag{24}$$

where $\tilde{\mathbf{a}}_j^t \sim \delta^t(\theta_j^*) = \left(\pi_j(H^0, \theta_j^*), ..., \pi_j(H^{t-1}, \theta_j^*)\right)$. The value of $p$ corresponds to the probability with which we expect to observe a test statistic at least as extreme as $T(\mathbf{a}_j^t, \hat{\mathbf{a}}_j^t)$, under the null-hypothesis that $\theta_j^* = \theta_j^+$. Thus, we reject $\theta_j^*$ if $p$ is below some "significance level" $\alpha^*$.

In the following subsections, we describe the test statistic T and its asymptotic properties, and how our algorithm learns the distribution of $T(\tilde{\mathbf{a}}_j^t, \hat{\mathbf{a}}_j^t)$. A summary of the algorithm is given in Algorithm 2.

#### 7.2.1. Test Statistic

We follow the general approach outlined earlier by which we compute a *score* from a vector of actions and their hypothesised distributions. Formally, we define a *score function* as $z : (A_j)^t \times \Delta(A_j)^t \to \mathbb{R}$, where $\Delta(A_j)$ is the set of all probability distributions over $A_j$. Thus, $z(\mathbf{a}_j^t, \delta^t(\theta_j^*))$ is the score for observed actions $\mathbf{a}_j^t$ and hypothesised distributions $\delta^t(\theta_j^*)$, and we sometimes abbreviate this to $z(\mathbf{a}_j^t, \theta_j^*)$. We use Z to denote the space of all score functions.

Given a score function $z$, we define the test statistic T as

$$T(\tilde{\mathbf{a}}_j^t, \hat{\mathbf{a}}_j^t) = \frac{1}{t} \sum_{\tau=1}^{t} T_\tau(\tilde{\mathbf{a}}_j^\tau, \hat{\mathbf{a}}_j^\tau) \tag{25}$$

$$T_\tau(\tilde{\mathbf{a}}_j^\tau, \hat{\mathbf{a}}_j^\tau) = z(\tilde{\mathbf{a}}_j^\tau, \theta_j^*) - z(\hat{\mathbf{a}}_j^\tau, \theta_j^*) \tag{26}$$

**Algorithm 2** Automatic behavioural hypothesis testing
---
    **Input:** history $H^t$ (including observed action $a_j^{t-1}$)
    **Output:** $p$-value (reject $\theta_j^*$ if $p$ below some threshold $\alpha^*$)
    **Parameters:** hypothesis $\theta_j^*$, score functions $z_1, ..., z_K$, $N > 0$
    // *Expand action vectors*
    Set $\mathbf{a}_j^t \leftarrow \langle \mathbf{a}_j^{t-1}, a_j^{t-1}\rangle$
    Sample $\hat{a}_j^{t-1} \sim \pi_j(H^{t-1}, \theta_j^*)$; set $\hat{\mathbf{a}}_j^t \leftarrow \langle \hat{\mathbf{a}}_j^{t-1}, \hat{a}_j^{t-1}\rangle$
    **for** $n = 1, ..., N$ **do**
        Sample $\tilde{a}_j^{t-1} \sim \pi_j(H^{t-1}, \theta_j^*)$; set $\tilde{\mathbf{a}}_j^{t,n} \leftarrow \langle \tilde{\mathbf{a}}_j^{t-1,n}, \tilde{a}_j^{t-1}\rangle$
    // *Fit skew-normal distribution f*
    **if** update parameters? **then**
        Compute $D \leftarrow \{\mathrm{T}(\tilde{\mathbf{a}}_j^{t,n}, \hat{\mathbf{a}}_j^t) \mid n = 1, ..., N\}$
        Fit $\xi, \omega, \beta$ to $D$, e.g. using (35)
        Find mode $\mu$ from $\xi, \omega, \beta$
    // *Compute p-value*
    Compute $q \leftarrow \mathrm{T}(\mathbf{a}_j^t, \hat{\mathbf{a}}_j^t)$ using (25)/(28)
    **return** $p \leftarrow f(q \mid \xi, \omega, \beta) / f(\mu \mid \xi, \omega, \beta)$
---

where $\tilde{\mathbf{a}}_j^\tau$ and $\hat{\mathbf{a}}_j^\tau$ denote the $\tau$-prefixes of $\tilde{\mathbf{a}}_j^t$ and $\hat{\mathbf{a}}_j^t$, respectively.

In this work, we assume that $z$ is provided by the user. While formally unnecessary (in the sense that our analysis does not require it), we find it a useful design guideline to interpret a score as a kind of likelihood, such that higher scores suggest higher likelihood of $\theta_j^*$ being correct. Under this interpretation, a minimum requirement for $z$ should be that it is *consistent*, such that, for any $t > 0$ and $\theta_j^* \in \Theta_j$,

$$\theta_j^* \in \Pi^z = \arg\max_{\theta_j' \in \Theta_j} \mathbb{E}_{\mathbf{a}_j' \sim \delta^t(\theta_j^*)} \left[ z(\mathbf{a}_j', \theta_j') \right] \tag{27}$$

where $\mathbb{E}_\eta$ denotes the expectation under $\eta$. This ensures that if the null-hypothesis $\theta_j^* = \theta_j^+$ is true, then the score $z(\mathbf{a}_j^t, \theta_j^*)$ is maximised on expectation.

Ideally, we would like a score function $z$ which is *perfect* in that it is consistent and $|\Pi^z| = 1$. This means that $\theta_j^*$ can maximise $z(\mathbf{a}_j^t, \theta_j^*)$ (where $\mathbf{a}_j^t \sim \delta^t(\theta_j^+)$) *only* if $\theta_j^* = \theta_j^+$. Unfortunately, it is unclear if such a score function exists for the general case and how it should look. Even if we restrict the behaviours agents may exhibit, it can still be difficult to find a perfect score function. On the other hand, it is a relatively simple task to specify a small set of score functions $z_1, ..., z_K$ which are consistent but imperfect. (Examples are given in Section 7.3.) Given that these score functions are consistent, we know that the cardinality $|\cap_k \Pi^{z_k}|$ can only monotonically decrease. Therefore, it seems a reasonable approach to combine multiple imperfect score functions in an attempt to approximate a perfect score function.

Given score functions $z_1, ..., z_K \in Z$ which are all bounded by the same interval $[a, b] \subset \mathbb{R}$, we redefine $\mathrm{T}_\tau$ to

$$\mathrm{T}_\tau(\tilde{\mathbf{a}}_j^\tau, \hat{\mathbf{a}}_j^\tau) = \sum_{k=1}^K w_k \left( z_k(\tilde{\mathbf{a}}_j^\tau, \theta_j^*) - z_k(\hat{\mathbf{a}}_j^\tau, \theta_j^*) \right) \tag{28}$$

where $w_k \in \mathbb{R}$ is a weight for score function $z_k$. In this work, we set $w_k = \frac{1}{K}$. (We also experiment with alternative weighting schemes in Section 7.3.) However, we believe that $w_k$ may serve as an interface for useful modifications of our algorithm. For example, Yue et al. (2010) compute weights to increase the power of their hypothesis tests.

### 7.2.2. Asymptotic Properties

The vectors $\mathbf{a}_j^t$ and $\hat{\mathbf{a}}_j^t$ are constructed iteratively. That is, at time $t$, we observe agent $j$'s past action $a_j^{t-1}$, which was generated from $\pi_j(H^{t-1}, \theta_j^+)$, and set $\mathbf{a}_j^t = \langle \mathbf{a}_j^{t-1}, a_j^{t-1} \rangle$. At the same time, we sample an action $\hat{a}_j^{t-1}$ using $\pi_j(H^{t-1}, \theta_j^*)$ and set $\hat{\mathbf{a}}_j^t = \langle \hat{\mathbf{a}}_j^{t-1}, \hat{a}_j^{t-1} \rangle$. Assuming the null-hypothesis $\theta_j^* = \theta_j^+$, will $T(\mathbf{a}_j^t, \hat{\mathbf{a}}_j^t)$ converge in the process?

Unfortunately, T might not converge. This may seem surprising at first glance given that $a_j^{t-1}$ and $\hat{a}_j^{t-1}$ have the same distribution $\pi_j(H^{t-1}, \theta_j^+) = \pi_j(H^{t-1}, \theta_j^*)$, since $\mathbb{E}_{x,y \sim \psi}[x - y] = 0$ for any distribution $\psi$. However, there is a subtle but important difference: while $a_j^{t-1}$ and $\hat{a}_j^{t-1}$ have the same distribution, $z_k(\mathbf{a}_j^t, \theta_j^*)$ and $z_k(\hat{\mathbf{a}}_j^t, \theta_j^*)$ may have arbitrarily different distributions. This is because these scores may depend on the entire prefix vectors $\mathbf{a}_j^{t-1}$ and $\hat{\mathbf{a}}_j^{t-1}$, respectively, which means that their distributions may be different if $\mathbf{a}_j^{t-1} \neq \hat{\mathbf{a}}_j^{t-1}$. Fortunately, our algorithm does not require T to converge because it learns the distribution of T during the interaction process, as we will discuss in Section 7.2.3.

Interestingly, while T may not converge, it can be shown that the fluctuation of T is eventually normally distributed, for any set of score functions $z_1, ..., z_K$ with bound $[a, b]$. Formally, let $\mathbb{E}[T_\tau(\mathbf{a}_j^\tau, \hat{\mathbf{a}}_j^\tau)]$ and $\text{Var}[T_\tau(\mathbf{a}_j^\tau, \hat{\mathbf{a}}_j^\tau)]$ denote the finite expectation and variance of $T_\tau(\mathbf{a}_j^\tau, \hat{\mathbf{a}}_j^\tau)$, where it is irrelevant if $\mathbf{a}_j^\tau, \hat{\mathbf{a}}_j^\tau$ are sampled directly from $\delta^\tau(\theta_j^*)$ or generated iteratively as prescribed above. Furthermore, let $\sigma_t^2 = \sum_{\tau=1}^t \text{Var}[T_\tau(\mathbf{a}_j^\tau, \hat{\mathbf{a}}_j^\tau)]$ denote the cumulative variance. Then, the standardised stochastic sum

$$\frac{1}{\sigma_t} \sum_{\tau=1}^{t} T_\tau(\mathbf{a}_j^\tau, \hat{\mathbf{a}}_j^\tau) - \mathbb{E}[T_\tau(\mathbf{a}_j^\tau, \hat{\mathbf{a}}_j^\tau)] \tag{29}$$

will converge in distribution to the standard normal distribution as $t \to \infty$. Thus, T is normally distributed as well.

To see this, first recall that the standard central limit theorem requires the random variables $T_\tau$ to be independent and identically distributed. In our case, $T_\tau$ are independent in that the random outcome of $T_\tau$ has no effect on the outcome of $T_{\tau'}$. However, $T_\tau$ and $T_{\tau'}$ depend on different action sequences, and may therefore have different distributions. Hence, we have to show an additional property, commonly known as *Lyapunov's condition* (e.g. Fischer, 2010), which states that there exists a positive integer $d$ such that

$$\lim_{t \to \infty} \frac{\hat{\sigma}_t^{2+d}}{\sigma_t^{2+d}} = 0, \text{ with} \tag{30}$$

$$\hat{\sigma}_t^{2+d} = \sum_{\tau=1}^{t} \mathbb{E}\left[\left|T_\tau(\mathbf{a}_j^\tau, \hat{\mathbf{a}}_j^\tau) - \mathbb{E}[T_\tau(\mathbf{a}_j^\tau, \hat{\mathbf{a}}_j^\tau)]\right|^{2+d}\right]. \tag{31}$$

Since $z_k$ are bounded, we know that $T_\tau$ are bounded. Hence, the summands in (31) are uniformly bounded, say by $U$ for brevity. Setting $d = 1$, we obtain

$$\lim_{t \to \infty} \frac{\hat{\sigma}_t^3}{\sigma_t^3} \leq \frac{U \hat{\sigma}_t^2}{\sigma_t^3} = \frac{U}{\sigma_t} \tag{32}$$

The last part goes to zero if $\sigma_t \to \infty$, and hence Lyapunov's condition holds. If, on the other hand, $\sigma_t$ converges, then this means that the variance of $T_\tau$ is zero from some point onward (or that it has an appropriate convergence to zero). From this point, $\theta_j^*$ prescribes fully deterministic action choices for agent $j$ (i.e. $\exists a_j : \pi_j(H^\tau, a_j, \theta_j^*) = 1$), and a statistical analysis is no longer necessary.

*7.2.3. Learning the Test Distribution*

Given that T is eventually normal, it may seem reasonable to compute (24) using a normal distribution whose parameters are fitted during the interaction. However, this fails to recognise that the distribution of T is shaped *gradually* over an extended time period, and that the fluctuation around T can be heavily skewed in either direction until convergence to a normal distribution emerges. Thus, a normal distribution may be a poor fit during this shaping period.

What is needed is a distribution which can represent any normal distribution, and which is flexible enough to faithfully represent the gradual shaping. One distribution which has these properties is the *skew-normal distribution* (Azzalini, 1985; O'Hagan and Leonard, 1976). Given the PDF $\phi$ and CDF $\Phi$ of the standard normal distribution, the skew-normal PDF is defined as

$$f(x \mid \xi, \omega, \beta) = \frac{2}{\omega} \phi\left(\frac{x - \xi}{\omega}\right) \Phi\left(\beta\left(\frac{x - \xi}{\omega}\right)\right) \tag{33}$$

where $\xi \in \mathbb{R}$ is the location parameter, $\omega \in \mathbb{R}^+$ is the scale parameter, and $\beta \in \mathbb{R}$ is the shape parameter. Note that this reduces to the normal PDF for $\beta = 0$, in which case $\xi$ and $\omega$ correspond to the mean and standard deviation, respectively. Hence, the normal distribution is a sub-class of the skew-normal distribution.

Our algorithm learns the shifting parameters of $f$ during the interaction process, using a simple but effective sampling procedure. Essentially, we use $\theta_j^*$ to iteratively generate $N$ additional action vectors $\tilde{\mathbf{a}}_j^{t,1}, ..., \tilde{\mathbf{a}}_j^{t,N}$ in the exact same way as $\hat{\mathbf{a}}_j^t$. The vectors $\tilde{\mathbf{a}}_j^{t,n}$ are then mapped into data points

$$D = \left\{ T(\tilde{\mathbf{a}}_j^{t,n}, \hat{\mathbf{a}}_j^t) \mid n = 1, ..., N \right\} \tag{34}$$

which are used to estimate the parameters $\xi, \omega, \beta$ by minimising the negative log-likelihood

$$N \log(\omega) - \sum_{x \in D} \log \phi\left(\frac{x - \xi}{\omega}\right) + \log \Phi\left(\beta\left(\frac{x - \xi}{\omega}\right)\right) \tag{35}$$

whilst ensuring that $\omega$ is positive. An alternative is the method-of-moments estimator, which can also be used to obtain initial values for (35). Note that it is usually unnecessary to estimate the parameters at every point in time; it seems reasonable to update the parameters less frequently as the amount of evidence (i.e. observed actions) grows.

Given the asymmetry of the skew-normal distribution, the semantics of "as extreme as" in (24) may no longer be obvious (e.g. is this with respect to the mean or mode?). In addition, the usual tail-area calculation of the *p*-value requires the CDF, but there is no closed form for the skew-normal CDF and approximating it is rather cumbersome. To circumvent these issues, we approximate the $p$-value as

$$p \approx \frac{f(T(\mathbf{a}_j^t, \hat{\mathbf{a}}_j^t) \mid \xi, \omega, \beta)}{f(\mu \mid \xi, \omega, \beta)} \quad (36)$$

where $\mu$ is the mode of the fitted skew-normal distribution. This avoids the asymmetry issue and is easier to compute.

### 7.3. Experiments

We conducted a comprehensive set of experiments to investigate the accuracy (correct and incorrect rejection), scalability (with number of actions), and sampling complexity of our algorithm. The following three score functions and their combinations were used:

$$z_1(\mathbf{a}_j^t, \theta_j^*) = \frac{1}{t} \sum_{\tau=0}^{t-1} \frac{\pi_j(H^\tau, a_j^\tau, \theta_j^*)}{\max_{a_j \in A_j} \pi_j(H^\tau, a_j, \theta_j^*)} \quad (37)$$

$$z_2(\mathbf{a}_j^t, \theta_j^*) = \frac{1}{t} \sum_{\tau=0}^{t-1} 1 - \mathbb{E}_{a_j \sim \pi_j(H^\tau, \theta_j^*)} \left| \pi_j(H^\tau, a_j^\tau, \theta_j^*) - \pi_j(H^\tau, a_j, \theta_j^*) \right| \quad (38)$$

$$z_3(\mathbf{a}_j^t, \theta_j^*) = \sum_{a_j \in A_j} \min \left[ \frac{1}{t} \sum_{\tau=0}^{t-1} [a_j^\tau = a_j]_1, \frac{1}{t} \sum_{\tau=0}^{t-1} \pi_j(H^\tau, a_j, \theta_j^*) \right] \quad (39)$$

where $[b]_1 = 1$ if $b$ is true and 0 otherwise. Note that $z_1, z_3$ are generally consistent (cf. Section 7.2.1), while $z_2$ is consistent for $|A_j| = 2$ but not necessarily for $|A_j| > 2$. Furthermore, $z_1, z_2, z_3$ are all imperfect. The score function $z_3$ is based on the empirical frequency distribution.

The parameters of the test distribution (cf. Section 7.2.3) were estimated less frequently as $t$ increased. The first estimation was performed at time $t = 1$ (i.e. after observing one action). After estimating the parameters at time $t$, we waited $\lfloor \sqrt{t} \rfloor - 1$ time steps until the parameters were re-fitted. Throughout our experiments, we used a significance level of $\alpha^* = 0.01$ (i.e. reject $\theta_j^*$ if the $p$-value is below 0.01).

#### 7.3.1. Random Behaviours

In the first set of experiments, the behaviour (type) spaces $\Theta_i$ and $\Theta_j$ were restricted to "random" behaviours. Each random behaviour is defined by a sequence of random probability distributions over $A_j$. The distributions are created by drawing uniform random numbers from $(0, 1)$ for each action $a_j \in A_j$, and subsequent normalisation so that the values sum up to 1.

Random behaviours are a good baseline for our experiments because they are usually hard to distinguish. This is due to the fact that the entire set $A_j$ is always in the support of the behaviours, and since they do not react to any past actions. These properties mean that there is little structure in the interaction that can be used to distinguish behaviours.

We simulated 1000 interaction processes, each lasting 10000 time steps. In each process, we randomly sampled behaviours $\theta_i \in \Theta_i, \theta_j^+ \in \Theta_j$ to control agents $i$ and $j$, respectively. In half of these processes, we used a correct hypothesis $\theta_j^* = \theta_j^+$. In the other half, we sampled a random hypothesis $\theta_j^* \in \Theta_j$ with $\theta_j^* \neq \theta_j^+$. We repeated each set of simulations for $|A_j| = 2, 10, 20$ (with $|A_i| = |A_j|$) and $N = 10, 50, 100$ (cf. Section 7.2.3).

Figure 6 shows the average accuracy of our algorithm (for $N = 50$), by which we mean the average percentage of time steps in which the algorithm made correct decisions (i.e. no reject if $\theta_j^* = \theta_j^+$; reject if $\theta_j^* \neq \theta_j^+$). The x-axis shows the combination of score functions used to compute the test statistic (e.g. [1 2] means that we combined $z_1, z_2$).

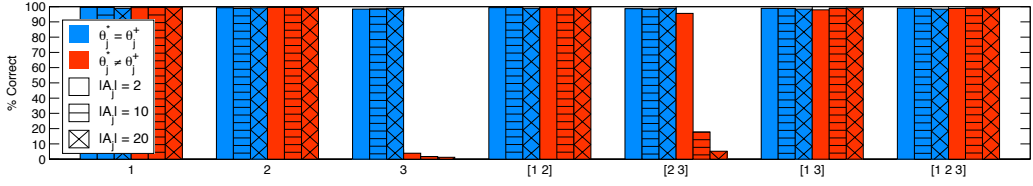

Figure 6: Average accuracy with random behaviours, for $N = 50$ and $|A_j| = 2, 10, 20$. Results averaged over 500 processes with 10000 time steps, for $\theta_j^* = \theta_j^+$ and $\theta_j^* \neq \theta_j^+$ each. X-axis shows score functions $z_k$ used in test statistic.

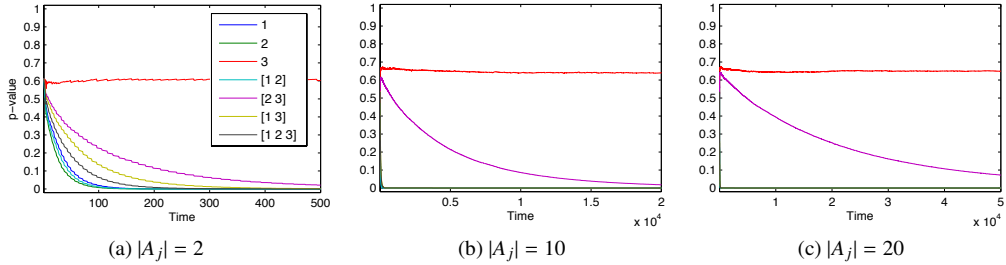

Figure 7: Average $p$-values with random behaviours, for $N = 50$ and $\theta_j^* \neq \theta_j^+$ (i.e. hypothesis wrong). Results averaged over 500 processes. Legend shows the score functions $z_k$ used in test statistic.

The results show that our algorithm achieved excellent accuracy, often bordering the 100% mark. They also show that the algorithm scaled well with the number of actions, with no degradation in accuracy. However, there were two exceptions to these observations: using $z_3$ resulted in very poor accuracy for $\theta_j^* \neq \theta_j^+$, and the combination of $z_2, z_3$ scaled badly for $\theta_j^* \neq \theta_j^+$.

The reason for both of these exceptions is that $z_3$ is not a good scoring scheme for random behaviours. The function $z_3$ quantifies a similarity between the empirical frequency distribution and the averaged hypothesised distributions. For random behaviours (as defined in this work), both of these distributions will converge to the uniform distribution. Thus, under $z_3$, any two random behaviours will eventually be the same, which explains the low accuracy for $\theta_j^* \neq \theta_j^+$.

As can be seen in Figure 6, the inadequacy of $z_3$ is solved when adding any of the other score functions $z_1, z_2$. These functions add discriminative information to the test statistic, which technically means that the cardinality $|\Pi^z|$ in (27) is reduced. However, in the case of $[z_2, z_3]$, the converge is substantially slower for higher $|A_j|$, meaning that more evidence is needed until $\theta_j^*$ can be rejected. Figure 7 shows how a higher number of actions affects the average convergence rate of $p$-values computed with $z_2, z_3$.

In addition to the score functions $z_k$, a central aspect for the convergence of $p$-values are the corresponding weights $w_k$ (cf. (28)). As mentioned in Section 7.2.1, we use uniform weights $w_k = \frac{1}{K}$. However, to show that the weighting is no trivial matter, we repeated our experiments with four alternative weighting schemes: Let $z_k^\tau = z_k(\tilde{\mathbf{a}}_j^\tau, \theta_j^*) - z_k(\hat{\mathbf{a}}_j^\tau, \theta_j^*)$ denote the summands in (28). The weighting schemes truemax/truemin assign $w_k = 1$ for the first $k$ that maximises/minimises $|z_k^\tau|$, and 0 otherwise. Similarly, the weighting schemes max/min assign $w_k = 1$ for the first $k$ that maximises/minimises $z_k^\tau$, and 0 otherwise.

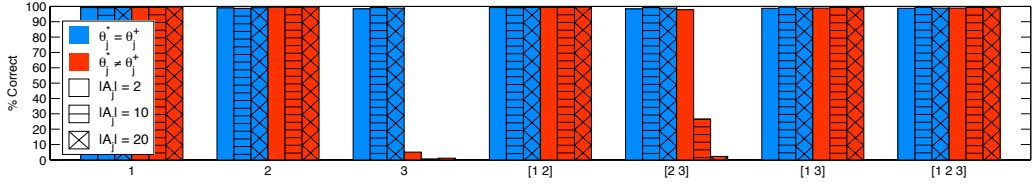

Figure 8: Average accuracy with random behaviours, for $N = 50$ and $|A_j| = 2, 10, 20$. Weights $w_k$ computed using `truemax` weighting. X-axis shows score functions $z_k$ used in test statistic.

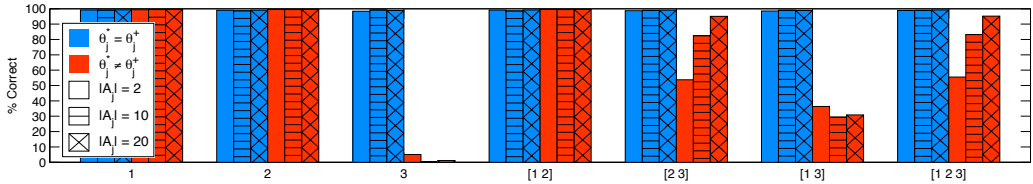

Figure 9: Average accuracy with random behaviours, for $N = 50$ and $|A_j| = 2, 10, 20$. Weights $w_k$ computed using `truemin` weighting. X-axis shows score functions $z_k$ used in test statistic.

Figures 8 and 9 show the results for `truemax` and `truemin`. As can be seen in the figures, `truemax` is very similar to uniform weights while `truemin` improves the convergence for $[z_2, z_3]$ but compromises elsewhere. The results for `max` and `min` are very similar to those of `truemin` and `truemax`, respectively, hence we omit them.

Finally, we recomputed all accuracies using a more lenient significance level of $\alpha^* = 0.05$. As could be expected, this marginally decreased and increased (i.e. by a few percentage points) the accuracy for $\theta_j^* = \theta_j^+$ and $\theta_j^* \neq \theta_j^+$, respectively. This was primarily observed in the early stages of the interaction. Overall, however, the results were very similar to those obtained with $\alpha^* = 0.01$.

Recall that $N$ specifies the number of sampled action vectors $\tilde{\mathbf{a}}_j^{t,n}$ used to learn the distribution of the test statistic (cf. Section 7.2.3). In the previous section, we reported results for $N = 50$. In this section, we investigate differences in accuracy for $N = 10, 50, 100$.

Figures 10 and 11 show the differences for $|A_j| = 2, 20$, respectively. (The figure for $|A_j| = 10$ was virtually the same as the one for $|A_j| = 20$, except with minor improvements in accuracy for the $[z_2, z_3]$ cluster. Hence, we omit it here.) As can be seen, there were improvements of up to 10% from $N = 10$ to $N = 50$, and no (or very marginal) improvements from $N = 50$ to $N = 100$. This was observed for all $|A_j| = 2, 10, 20$, and all constellations of score functions. The fact that $N = 50$ was sufficient even for $|A_j| = 20$ is remarkable, since, under random behaviours, there are $20^t$ possible action vectors to sample at any time $t$.

We also compared the learned skew-normal distributions and found that they fitted the data very well. Figures 12 and 13 show the histograms and fitted skew-normal distributions for two example processes after 1000 time steps. In Figure 13, we deliberately chose an example in which the learned distribution was maximally skewed for $N = 10$, which is a sign that $N$ was too small. Nonetheless, in the majority of the processes, the learned distribution was only moderately skewed and our algorithm achieved an average accuracy of 90% even for $N = 10$. Moreover, if one wants to avoid maximally skewed distributions, one can simply restrict the parameter space when fitting the skew-normal (specifically, the shape parameter $\beta$; cf. Section 7.2.3).

The flexibility of the skew-normal distribution was particularly useful in the early stages of the interaction, in which the test statistic typically does not follow a normal distribution.

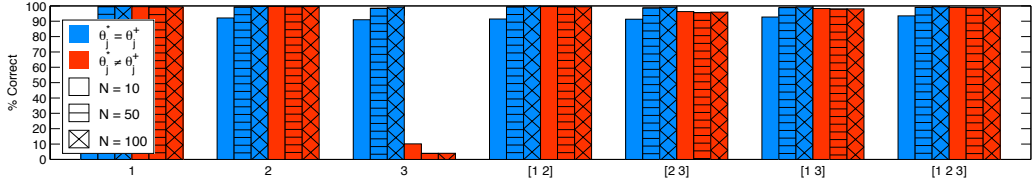

Figure 10: Average accuracy with random behaviours, for $|A_j| = 2$ and $N = 10, 50, 100$. Results averaged over 500 processes with 10000 time steps, for $\theta_j^* = \theta_j^+$ and $\theta_j^* \neq \theta_j^+$ each. X-axis shows score functions $z_k$ used in test statistic.

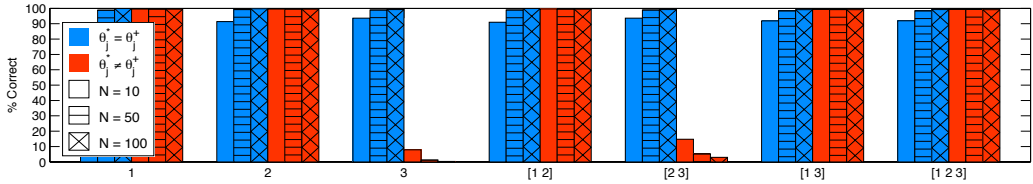

Figure 11: Average accuracy with random behaviours, for $|A_j| = 20$ and $N = 10, 50, 100$. Results averaged over 500 processes with 10000 time steps, for $\theta_j^* = \theta_j^+$ and $\theta_j^* \neq \theta_j^+$ each. X-axis shows score functions $z_k$ used in test statistic.

Figure 14 shows the test distribution for an example process after 10 time steps, using $z_2$ for the test statistic and $N = 100$ (the histogram was created using $N = 10000$). The learned skew-normal approximated the true test distribution very closely. Note that, in such examples, the normal and Student distributions do not produce good fits.

Our implementation of the algorithm performed all calculations as iterative updates (except for the skew-normal fitting). Hence, it used little (fixed) memory and had very low computation times. For example, using all three score functions and $|A_j| = 20, N = 100$, one cycle in the algorithm (cf. Algorithm 2) took on average less than 1 millisecond without fitting the skew-normal parameters, and less than 10 milliseconds when fitting the skew-normal parameters (using an off-the-shelf Simplex-optimiser with default parameters). The times were measured using Matlab R2014a on a Unix machine with a 2.6 GHz Intel Core i5 processor.

*7.3.2. Adaptive Behaviours*

We complemented the "structure-free" interaction of random behaviours by conducting analogous experiments with three additional classes of behaviours. Specifically, we used the benchmark framework specified in Section 5, which consists of 78 distinct $2 \times 2$ matrix games and three methods to automatically generate sets of behaviours for any given game. The three behaviour classes are Leader-Follower-Trigger Agents (LFT), Co-Evolved Decision Trees (CDT), and Co-Evolved Neural Networks (CNN). These classes cover a broad spectrum of possible behaviours, including fully deterministic (CDT), fully stochastic (CNN), and hybrid (LFT) behaviours. Furthermore, all generated behaviours are *adaptive* to varying degrees (i.e. they adapt their action choices based on the other player's choices). Detailed descriptions of the games and behaviour classes can be found in the appendix of (Albrecht, 2015).

The following experiments were performed for each behaviour class, using identical randomisation: For each of the 78 games, we simulated 10 interaction processes, each lasting 10000 time steps. For each process, we randomly sampled behaviours $\theta_i \in \Theta_i, \theta_j^+ \in \Theta_j$ to control agents $i$ and $j$, respectively, where $\Theta_i, \Theta_j$ were restricted to the same behaviour class. In half of these processes, we used a correct hypothesis $\theta_j^* = \theta_j^+$, and in the other half, we sampled a random hy-pothesis $\theta_j^* \in \Theta_j$ with $\theta_j^* \neq \theta_j^+$. As before, we repeated each simulation for $N = 10, 50, 100$ and all constellations of score functions, but found that there were virtually no differences. Hence, in the following, we report results for $N = 50$ and the $[z_1, z_2, z_3]$ cluster.

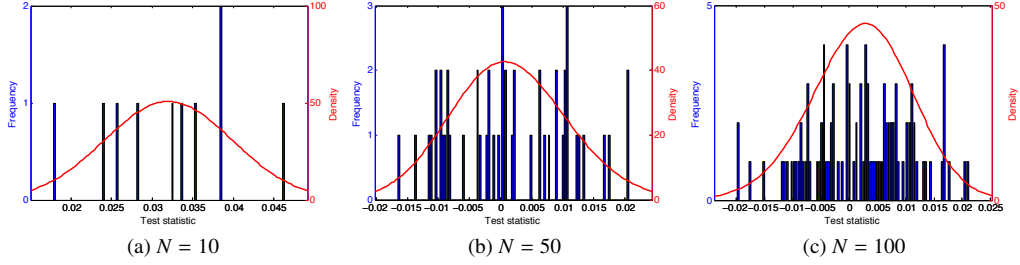

Figure 12: Example histograms and fitted skew-normal distributions (shown in red curve) after 1000 time steps, for random behaviours with $|A_j| = 10$ and $N = 10, 50, 100$. Using score function $z_1$ in test statistic.

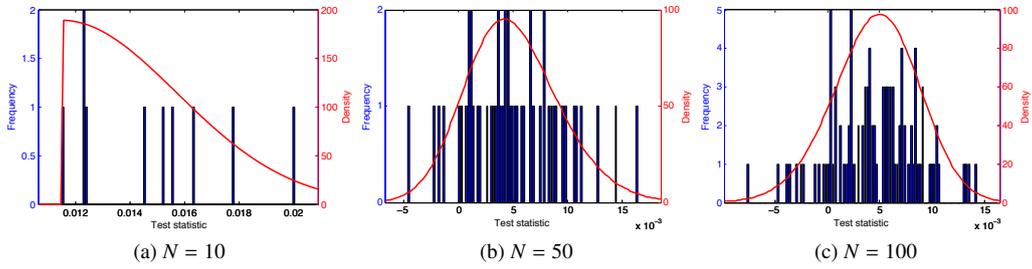

Figure 13: Example histograms and fitted skew-normal distributions (shown in red curve) after 1000 time steps, for random behaviours with $|A_j| = 10$ and $N = 10, 50, 100$. Using score functions $z_1, z_2, z_3$ in test statistic.

Figure 15a shows the average accuracy achieved by our algorithm for all three behaviour classes. While the accuracy for $\theta_j^* = \theta_j^+$ was generally good, the accuracy for $\theta_j^* \neq \theta_j^+$ was mixed. Note that this was not merely due to the fact that the score functions were imperfect (cf. Section 7.2.1), since we obtained the same results for all combinations. Rather, this reveals an inherent limitation of our approach, which is that *we do not actively probe aspects of the hypothesis $\theta_j^*$*. In other words, our algorithm performs statistical hypothesis tests based only on evidence that was generated by $\theta_i$.

To illustrate this, it is useful to consider the tree structure of behaviours in the CDT class. Each node in a tree $\theta_j^+$ corresponds to a past action taken by $\theta_i$. Depending on how $\theta_i$ chooses actions, we may only ever see a subset of the entire tree that defines $\theta_j^+$. However, if our hypothesis $\theta_j^*$ differs from $\theta_j^+$ only in the unseen aspects of $\theta_j^+$, then there is no way for our algorithm to differentiate the two. Hence the asymmetry in accuracy for $\theta_j^* = \theta_j^+$ and $\theta_j^* \neq \theta_j^+$. Note that this problem did not occur in random behaviours because, there, all aspects are eventually visible.

Following this observation, we repeated the same experiments but restricted $\Theta_i$ to random behaviours (cf. Section 7.3.1), with the goal of exploring $\theta_j^*$ more thoroughly. As shown in Figure 15b, this led to significant improvements in accuracy, especially for the CDT class. Nonetheless, choosing actions purely randomly may not be a sufficient probing strategy, hence the accuracy for CNN was still relatively low. For CNN, this was further complicated by the fact that two neural net-works $\theta_j, \theta_j'$ may formally be different ($\theta_j \neq \theta_j'$) but have essentially the same action probabilities (with extremely small differences). Hence, in such cases, we would require much more evidence to distinguish the behaviours.

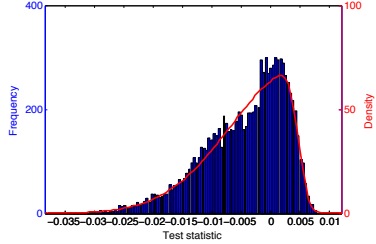

Figure 14: Example of true test distribution for $z_2$ and learned skew-normal distribution (shown in red curve) after 10 time steps, with $|A_j| = 10$ and $N = 100$.

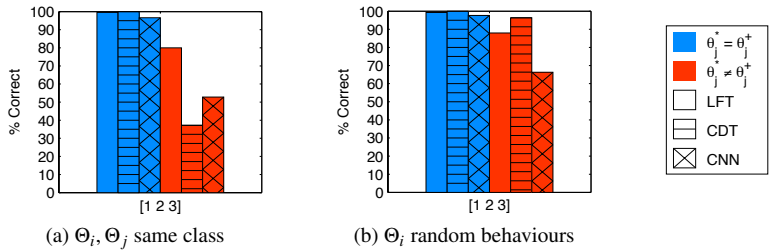

(a) $\Theta_i, \Theta_j$ same class

(b) $\Theta_i$ random behaviours

Figure 15: Average accuracy with behaviour classes LFT, CDT, CNN, for $N = 50$. Results averaged over 500 processes with 10000 time steps, for $\theta_j^* = \theta_j^+$ and $\theta_j^* \neq \theta_j^+$ each. Bars shown for $[z_1, z_2, z_3]$ test statistic.

## 8. Conclusion

Much work in artificial intelligence is focused on innovative applications such as adaptive user interfaces, robotic elderly care, and automated trading agents. A key technological challenge in these applications is to design an intelligent agent which can quickly learn to interact effectively with other agents whose behaviours are initially unknown. Learning from scratch in such problems is not a viable solution, since time is a crucial factor and exploration via trial-and-error may not be feasible or desirable. Instead, it is likely that any solution to this problem will have to draw heavily on prior experience and intuition, such as in the form of hypothesised behaviours. Indeed, if we have a strong intuition regarding the behaviour of other agents, e.g. based on past experience or structural constraints of the task to be completed, then this intuition should be utilised in the interaction. This is the motivation behind the type-based method studied in this work.

The idea in the type-based method is to hypothesise a set of possible behaviours, or types, which the other agents might have, and to plan our own actions with respect to those types which we believe are most likely, given the observed actions of the agents. In this regard, we identified and addressed a spectrum of important questions, pertaining to properties of beliefs over types and the possibility of incorrect types. Specifically, we formulated three alternative methods to incorporate observations into beliefs and studied the conditions under which the resulting beliefs will be correct or incorrect. We then investigated the impact of prior beliefs on payoff maximisation and methods to automatically compute prior beliefs. For the case in which our hypothesised types are incorrect, we analysed the conditions under which we are nevertheless able to complete our task, despite the incorrectness of types. Finally, we described an automatic statistical analysis which can be used to ascertain the correctness of hypothesised types during the interaction.

In addition to the theoretical insights, the results presented in this article have a number of practical implications: First of all, our analysis in Section 4 shows that the standard posterior formulation, in which the likelihood is defined as a product of action probabilities, may not always be an appropriate choice. Rather, one should also consider alternative formulations for posterior beliefs, such as the sum or correlated posteriors. Furthermore, our empirical analysis in Section 5 shows that prior beliefs can be crucial to our ability to maximise payoffs in the long-term. Indeed, we can often do better than a conservative uniform prior belief, by using automatic methods such as the ones used in this work. Another important practical implication is that we can use efficient model checking methods to verify optimality of hypothesised types. Specifically, in Section 6, we show a useful connection to probabilistic bisimulation checking. Moreover, for the case in which a prior analysis based on bisimulation is not possible, we show that the correctness of types can still be contemplated during the interaction. Our algorithm in Section 7 is simple to implement, highly efficient, and achieves high accuracy and scalability.

There are several potential directions for future work: Further formulations of posterior beliefs could be developed, and it would be interesting to know if the asymptotic correctness analysis in Section 4 could be complemented by useful finite-time error bounds. Our empirical analysis of prior beliefs in Section 5 could be refined by a theoretical analysis, and an important question is if prior beliefs can be computed with useful error bounds (the LP-priors are a step in this direction). Furthermore, the optimality analysis in Section 6 is focused on task completion and could be extended by an analysis focusing on payoff maximisation. Finally, it is unclear if the concept of perfect scores in Section 7 is generally feasible or even necessary, and what impact score weights have on convergence and decision quality.

Two aspects which we did not address, yet which are crucial to a successful deployment of the type-based method, are the complexity of the planning step and the size of the hypothesised type spaces. Regarding the former, it can be seen in Algorithm 1 (specifically (3)/(4)) that the time complexity of the planning is exponential in factors such as the number of agents, actions, and states, making it a very costly operation in complex systems. A promising solution are stochastic sampling procedures such as those used in (Albrecht and Ramamoorthy, 2013a; Barrett et al., 2011). Regarding the latter, the problem is that the number of types one may wish to hypothesise can grow dramatically with the size of the interaction problem (e.g. states, actions, agents). This is problematic because the predictions of each type must be computed at each point in time, hence it is desirable to minimise the number of hypothesised types. One way to do so is to develop methods which can produce small sets of reasonable types with good coverage of behaviours, in the spirit of works such as (Crandall, 2015). Another method would be to introduce learnable structure in types (i.e. parameters) such that each type covers a spectrum of behaviours. However, this would require an ability to infer the parameters from the interaction history.

**Acknowledgements:** The authors acknowledge the support of the German National Academic Foundation, the Masdar Institute-MIT collaborative agreement (Flagship Project 13CAMA1), the UK Engineering and Physical Sciences Research Council (EP/H012338/1), the European Commission (TOMSY Grant 270436, FP7-ICT-2009.2.1 Call 6), the Royal Academy of Engineering (Ingenious Grant), and the European Commission through SmartSociety Grant agreement no. 600854 (FOCAS ICT-2011.9.10). The authors wish to thank anonymous reviewers for their comments and suggestions.

**Appendix A. Proof of Theorem 1**

*Proof.* Kalai and Lehrer (1993) studied a model which can be equivalently described as a single-state SBG (i.e. $|S| = 1$) with a pure type distribution and product posterior. They showed that, if the player's assessment of future play is *absolutely continuous* with respect to the true probabilities of future play (i.e. any event that has true positive probability is assigned positive probability by the player), then (6) must hold. In our case, absolute continuity always holds by Assumption 1 and the fact that the prior probabilities $P_j$ are positive as well as the fact that the type distribution is pure, from which we can infer that the true types always have positive posterior probability.

In this proof, we seek to extend the convergence result of Kalai and Lehrer (1993) (henceforth KL) to multi-state SBGs with pure type distributions. Our strategy is to translate a SBG $\Gamma$ into a *modified SBG* $\hat{\Gamma}$ which is equivalent to $\Gamma$ in the sense that the players behave identically, and which is compatible to the model used in KL in the sense that the informational assumptions therein ignore the differences. We achieve this by introducing a new player *nature*, denoted $\xi$, which emulates the transitions of $\Gamma$ in $\hat{\Gamma}$.

Given a SBG $\Gamma = (S, s^0, \bar{S}, N, A_i, \Theta_i, u_i, \pi_i, T, \Upsilon)$, we define the modified SBG $\hat{\Gamma}$ as follows: Firstly, $\hat{\Gamma}$ has only one state, which can be arbitrary since it has no effect. The players in $\hat{\Gamma}$ are $\hat{N} = N \cup \{\xi\}$ where $i \in N$ have the same actions and types as in $\Gamma$ (i.e. $A_i$ and $\Theta_i$), and where we define the actions and types of $\xi$ to be $A_\xi = \Theta_\xi = S$ (i.e. nature's actions and types correspond to the states of $\Gamma$). The payoffs of $\xi$ are always zero and the strategy of $\xi$ at time $t$ is defined as

$$\pi_\xi^t(H^\tau, a_\xi, \theta_\xi) = \begin{cases} 0 & \tau = t, a_\xi \not\equiv \theta_\xi \\ 1 & \tau = t, a_\xi \equiv \theta_\xi \\ T(a_\xi^{\tau-1}, (a_i^{\tau-1})_{i \in N}, a_\xi) & \tau > t \end{cases}$$

where $H^\tau$ is any history of length $\tau \geq t$. ($H^\tau$ allows the players $i \in N$ to use $\pi_\xi^t$ for future predictions about $\xi$'s actions. This will be necessary to establish equivalence of $\hat{\Gamma}$ and $\Gamma$.)

The purpose of $\xi$ is to emulate the state transitions of $\Gamma$. Therefore, the modified strategies $\hat{\pi}_i$ and payoffs $\hat{u}_i$ of $i \in N$ are now defined with respect to the actions and types (since the current type of $\xi$ determines its next action) of $\xi$. Formally, $\hat{\pi}_i(H^t, a_i, \theta_i) = \pi_i(\bar{H}^t, a_i, \theta_i)$ where

$$\bar{H}^t = (\theta_\xi^0, (a_i^0)_{i \in N}, \theta_\xi^1, (a_i^1)_{i \in N}, ..., \theta_\xi^t)$$

and $\hat{u}_i(s, a^t, \theta_i^t) = u_i(\theta_\xi^t, (a_j^t)_{j \in N}, \theta_i^t)$, where $s$ is the only state of $\hat{\Gamma}$ and $a^t \in \times_{i \in \hat{N}} A_i$.

Finally, $\hat{\Gamma}$ uses two type distributions, $\Upsilon$ and $\Upsilon_\xi$, where $\Upsilon$ is the type distribution of $\Gamma$ and $\Upsilon_\xi$ is defined as $\Upsilon_\xi(H^t, \theta_\xi) = T(a_\xi^{t-1}, (a_i^{t-1})_{i \in N}, \theta_\xi)$. If $s^0$ is the initial state of $\Gamma$, then $\Upsilon_\xi(H^0, \theta_\xi) = 1$ for $\theta_\xi \equiv s^0$.

The modified SBG $\hat{\Gamma}$ proceeds as the original SBG $\Gamma$, except for the following changes: *(a)* $\Upsilon$ is used to sample the types for $i \in N$ (as usual) while $\Upsilon_\xi$ is used to sample the types for $\xi$; *(b)* each player is informed about its own type *and* the type of $\xi$. This completes the definition of $\hat{\Gamma}$.

The modified SBG $\hat{\Gamma}$ is equivalent to the original SBG $\Gamma$ in the sense that the players $i \in N$ have identical behaviour in both SBGs. Since the players always know the type of $\xi$, they also know the next action of $\xi$, which corresponds to knowing the current state of the game. Furthermore, note that the strategy of $\xi$ uses two time indices, $t$ and $\tau$, which allow it to distinguish between the current time ($\tau = t$) and a future time ($\tau > t$). This means that $\pi_\xi^t$ can be used to compute expected payoffs in $\hat{\Gamma}$ in the same way as $T$ is used to compute expected payoffs in $\Gamma$. In other words, the formulas (2) and (3) can be modified in a straightforward manner by replacing the original components of Γ with the modified components of $\hat{Γ}$, yielding the same results. Finally, since $\hat{Γ}$ uses the same type distribution as Γ to sample types for $i \in N$, there are no differences in their payoffs and strategies.

To complete the proof, we note that *(a)* and *(b)* are the only procedural differences between the modified SBG and the model used in KL. However, since we specify that the players always know the type of $ξ$, there is no need to learn the type distribution $Υ_ξ$, hence *(a)* and *(b)* have no effect in KL. The important point is that KL assume a model in which the players only interact with other players, but not with an environment. Since we eliminated the environment by replacing it with a player $ξ$, this is precisely what happens in the modified SBG. Therefore, the convergence result of KL carries over to multi-state SBGs with pure type distributions. □

## References

Albrecht, S., 2015. Utilising policy types for effective ad hoc coordination in multiagent systems. Ph.D. thesis, The University of Edinburgh.

Albrecht, S., Crandall, J., Ramamoorthy, S., 2015. An empirical study on the practical impact of prior beliefs over policy types. In: Proceedings of the 29th AAAI Conference on Artificial Intelligence. pp. 1988–1994.

Albrecht, S., Ramamoorthy, S., 2012. Comparative evaluation of MAL algorithms in a diverse set of ad hoc team problems. In: Proceedings of the 11th International Conference on Autonomous Agents and Multiagent Systems. pp. 349–356.

Albrecht, S., Ramamoorthy, S., 2013a. A game-theoretic model and best-response learning method for ad hoc coordination in multiagent systems. Tech. rep., School of Informatics, The University of Edinburgh.
URL http://arxiv.org/abs/1506.01170

Albrecht, S., Ramamoorthy, S., 2013b. A game-theoretic model and best-response learning method for ad hoc coordination in multiagent systems (extended abstract). In: Proceedings of the 12th International Conference on Autonomous Agents and Multiagent Systems. pp. 1155–1156.

Albrecht, S., Ramamoorthy, S., 2014. On convergence and optimality of best-response learning with policy types in multiagent systems. In: Proceedings of the 30th Conference on Uncertainty in Artificial Intelligence. pp. 12–21.

Albrecht, S., Ramamoorthy, S., 2015. Are you doing what I think you are doing? Criticising uncertain agent models. In: Proceedings of the 31st Conference on Uncertainty in Artificial Intelligence. pp. 52–61.

Aumann, R., 1974. Subjectivity and correlation in randomized strategies. Journal of mathematical Economics 1, 67–96.

Azzalini, A., 1985. A class of distributions which includes the normal ones. Scandinavian Journal of Statistics 12, 171–178.

Baier, C., 1996. Polynomial time algorithms for testing probabilistic bisimulation and simulation. In: Proceedings of the 8th International Conference on Computer Aided Verification, Lecture Notes in Computer Science. Vol. 1102. Springer, pp. 38–49.

Barrett, S., Stone, P., Kraus, S., 2011. Empirical evaluation of ad hoc teamwork in the pursuit domain. In: Proceedings of the 10th International Conference on Autonomous Agents and Multiagent Systems. pp. 567–574.

Barrett, S., Stone, P., Kraus, S., Rosenfeld, A., 2013. Teamwork with limited knowledge of teammates. In: Proceedings of the 27th AAAI Conference on Artificial Intelligence. pp. 102–108.

Basawa, I., Scott, D., 1977. Efficient tests for stochastic processes. Sankhyā: The Indian Journal of Statistics, Series A, 21–31.

Bayarri, M., Berger, J., 2000. *P* values for composite null models. Journal of the American Statistical Association 95 (452), 1127–1142.

Bellman, R., 1957. Dynamic Programming. Princeton University Press.

Berger, J., Sellke, T., 1987. Testing a point null hypothesis: the irreconcilability of *P* values and evidence (with discussion). Journal of the American Statistical Association 82, 112–122.

Bernardo, J., 1979. Reference posterior distributions for Bayesian inference. Journal of the Royal Statistical Society. Series B (Methodological) 41 (2), 113–147.

Bernstein, D., Zilberstein, S., Immerman, N., 2000. The complexity of decentralized control of Markov decision processes. In: Proceedings of the 16th Conference on Uncertainty in Artificial Intelligence. pp. 32–37.

Bowling, M., McCracken, P., 2005. Coordination and adaptation in impromptu teams. In: Proceedings of the 20th National Conference on Artificial Intelligence. pp. 53–58.

Bowling, M., Veloso, M., 2002. Multiagent learning using a variable learning rate. Artificial Intelligence 136 (2), 215–250.

Box, G., 1980. Sampling and Bayes' inference in scientific modelling and robustness. Journal of the Royal Statistical Society. Series A (General), 383–430.

Brafman, R., Tennenholtz, M., 2003. R-max – A general polynomial time algorithm for near-optimal reinforcement learning. Journal of Machine Learning Research 3, 213–231.

Brown, G., 1951. Iterative solution of games by fictitious play. Activity Analysis of Production and Allocation 13 (1), 374–376.

Carberry, S., 2001. Techniques for plan recognition. User Modeling and User-Adapted Interaction 11 (1-2), 31–48.

Carmel, D., Markovitch, S., 1999. Exploration strategies for model-based learning in multi-agent systems: exploration strategies. Autonomous Agents and Multi-Agent Systems 2 (2), 141–172.

Chalkiadakis, G., Boutilier, C., 2003. Coordination in multiagent reinforcement learning: a Bayesian approach. In: Proceedings of the 2nd International Conference on Autonomous Agents and Multiagent Systems. pp. 709–716.

Charniak, E., Goldman, R., 1993. A Bayesian model of plan recognition. Artificial Intelligence 64 (1), 53–79.

Conitzer, V., Sandholm, T., 2007. AWESOME: a general multiagent learning algorithm that converges in self-play and learns a best response against stationary opponents. Machine Learning 67 (1-2), 23–43.

Conitzer, V., Sandholm, T., 2008. New complexity results about Nash equilibria. Games and Economic Behavior 63 (2), 621–641.

Cox, D., 1977. The role of significance tests (with discussion). Scandinavian Journal of Statistics 4, 49–70.

Crandall, J., 2014. Towards minimizing disappointment in repeated games. Journal of Artificial Intelligence Research 49, 111–142.

Crandall, J., 2015. Robust learning in repeated stochastic games using meta-gaming. In: Proceedings of the 24th International Joint Conference on Artificial Intelligence. pp. 3416—3422.

De Finetti, B., 2008. Philosophical Lectures on Probability: Collected, Edited, and Annotated by Alberto Mura. Springer.

Dearden, R., Friedman, N., Andre, D., 1999. Model based Bayesian exploration. In: Proceedings of the 15th Conference on Uncertainty in Artificial Intelligence. pp. 150–159.

Dekel, E., Fudenberg, D., Levine, D., 2004. Learning to play Bayesian games. Games and Economic Behavior 46 (2), 282–303.

Dibangoye, J., Amato, C., Doniec, A., Charpillet, F., 2013. Producing efficient error-bounded solutions for transition independent decentralized MDPs. In: Proceedings of the 12th International Conference on Autonomous Agents and Multiagent Systems. pp. 539–546.

Doshi, P., Gmytrasiewicz, P., 2006. On the difficulty of achieving equilibrium in interactive POMDPs. In: Proceedings of the 21st National Conference on Artificial Intelligence. pp. 1131–1136.

Doshi, P., Gmytrasiewicz, P., 2009. Monte carlo sampling methods for approximating interactive POMDPs. Journal of Artificial Intelligence Research, 297–337.

Doshi, P., Perez, D., 2008. Generalized point based value iteration for interactive POMDPs. In: Proceedings of the 23rd AAAI Conference on Artificial Intelligence. pp. 63–68.

Doshi, P., Qu, X., Goodie, A., Young, D., 2010. Modeling recursive reasoning by humans using empirically informed interactive POMDPs. In: Proceedings of the 9th International Conference on Autonomous Agents and Multiagent Systems. pp. 1223–1230.

Doshi, P., Zeng, Y., Chen, Q., 2009. Graphical models for interactive POMDPs: representations and solutions. Autonomous Agents and Multi-Agent Systems 18 (3), 376–416.

Emery-Montemerlo, R., Gordon, G., Schneider, J., Thrun, S., 2004. Approximate solutions for partially observable stochastic games with common payoffs. In: Proceedings of the 3rd International Conference on Autonomous Agents and Multiagent Systems. pp. 136–143.

Etessami, K., Yannakakis, M., 2010. On the complexity of Nash equilibria and other fixed points. SIAM Journal on Computing 39 (6), 2531–2597.

Fischer, H., 2010. A History of the Central Limit Theorem: From Classical to Modern Probability Theory. Springer Science & Business Media.

Fisher, R., 1935. The Design of Experiments. Oliver & Boyd.

Foster, D., Young, H., 2001. On the impossibility of predicting the behavior of rational agents. Proceedings of the National Academy of Sciences 98 (22), 12848–12853.

Foster, D., Young, H., 2003. Learning, hypothesis testing, and Nash equilibrium. Games and Economic Behavior 45 (1), 73–96.

Fudenberg, D., Levine, D., 1993. Self-confirming equilibrium. Econometrica, 523–545.

Gal, Y., Pfeffer, A., Marzo, F., Grosz, B., 2004. Learning social preferences in games. In: Proceedings of the 19th National Conference on Artificial Intelligence. pp. 226–231.

Geib, C., Goldman, R., 2009. A probabilistic plan recognition algorithm based on plan tree grammars. Artificial Intelligence 173 (11), 1101–1132.

Gelman, A., Shalizi, C., 2013. Philosophy and the practice of Bayesian statistics. British Journal of Mathematical and Statistical Psychology 66 (1), 8–38.

Gmytrasiewicz, P., Doshi, P., 2005. A framework for sequential planning in multiagent settings. Journal of Artificial Intelligence Research 24 (1), 49–79.

Grosz, B., Kraus, S., 1996. Collaborative plans for complex group action. Artificial Intelligence 86 (2), 269–357.

Hansen, E., Bernstein, D., Zilberstein, S., 2004. Dynamic programming for partially observable stochastic games. In:

Proceedings of the 19th National Conference on Artificial Intelligence. pp. 709–715.

Hansson, H., Jonsson, B., 1994. A logic for reasoning about time and reliability. Formal Aspects of Computing 6 (5), 512–535.

Harsanyi, J., 1967. Games with incomplete information played by "Bayesian" players. Part I. The basic model. Management Science 14 (3), 159–182.

Harsanyi, J., 1968a. Games with incomplete information played by "Bayesian" players. Part II. Bayesian equilibrium points. Management Science 14 (5), 320–334.

Harsanyi, J., 1968b. Games with incomplete information played by "Bayesian" players. Part III. The basic probability distribution of the game. Management Science 14 (7), 486–502.

Holland, J., 1975. Adaptation in Natural and Artificial Systems: An Introductory Analysis with Applications to Biology, Control, and Artificial Intelligence. The MIT Press.

Howard, R., 1966. Information value theory. IEEE Transactions on Systems Science and Cybernetics 2 (1), 22–26.

Hu, J., Wellman, M., 2003. Nash q-learning for general-sum stochastic games. The Journal of Machine Learning Research 4, 1039–1069.

Jaynes, E., 1968. Prior probabilities. IEEE Transactions on Systems Science and Cybernetics 4 (3), 227–241.

Jordan, J., 1991. Bayesian learning in normal form games. Games and Economic Behavior 3 (1), 60–81.

Kahneman, D., Tversky, A., 1979. Prospect theory: an analysis of decision under risk. Econometrica 47, 263–292.

Kalai, E., Lehrer, E., 1993. Rational learning leads to Nash equilibrium. Econometrica 61 (5), 1019–1045.

Kaminka, G., Frenkel, I., 2007. Integration of coordination mechanisms in the BITE multi-robot architecture. In: Proceedings of the International Conference on Robotics and Automation. pp. 2859–2866.

Koza, J., 1992. Genetic Programming: On the Programming of Computers by Means of Natural Selection. The MIT Press.

Kuhn, H., 1953. Extensive games and the problem of information. Contributions to the Theory of Games 2 (28), 193–216.

Larsen, K., Skou, A., 1991. Bisimulation through probabilistic testing. Information and Computation 94 (1), 1–28.

Littman, M., 1994. Markov games as a framework for multi-agent reinforcement learning. In: Proceedings of the 11th International Conference on Machine Learning. Vol. 157. pp. 157–163.

Meng, X.-L., 1994. Posterior predictive $p$-values. The Annals of Statistics, 1142–1160.

Nachbar, J., 1997. Prediction, optimization, and learning in repeated games. Econometrica 65 (2), 275–309.

Nachbar, J., 2005. Beliefs in repeated games. Econometrica 73 (2), 459–480.

Nash, J., 1950. Equilibrium points in n-person games. Proceedings of the National Academy of Sciences 36 (1), 48–49.

Ng, B., Meyers, C., Boakye, K., Nitao, J., 2010. Towards applying interactive POMDPs to real-world adversary modeling. In: Proceedings of the 22nd Innovative Applications of Artificial Intelligence Conference. pp. 1814–1820.

Nyarko, Y., 1998. Bayesian learning and convergence to Nash equilibria without common priors. Economic Theory 11 (3), 643–655.

O'Hagan, A., Leonard, T., 1976. Bayes estimation subject to uncertainty about parameter constraints. Biometrika 63 (1), 201–203.

Rapoport, A., Guyer, M., 1966. A taxonomy of $2 \times 2$ games. General Systems: Yearbook of the Society for General Systems Research 11, 203–214.

Rubin, D., 1984. Bayesianly justifiable and relevant frequency calculations for the applied statistician. The Annals of Statistics 12 (4), 1151–1172.

Ryabko, D., Ryabko, B., 2008. On hypotheses testing for ergodic processes. In: Proceedings of IEEE Information Theory Workshop. pp. 281–283.

Shapley, L., 1953. Stochastic games. Proceedings of the National Academy of Sciences of the United States of America 39 (10), 1095.

Southey, F., Bowling, M., Larson, B., Piccione, C., Burch, N., Billings, D., Rayner, C., 2005. Bayes' bluff: opponent modelling in poker. In: Proceedings of the 21st Conference on Uncertainty in Artificial Intelligence. pp. 550–558.

Stone, P., Kaminka, G., Kraus, S., Rosenschein, J., 2010. Ad hoc autonomous agent teams: collaboration without pre-coordination. In: Proceedings of the 24th AAAI Conference on Artificial Intelligence. pp. 1504–1509.

Sukthankar, G., Goldman, R., Geib, C., Pynadath, D., Bui, H., 2014. Plan, Activity, and Intent Recognition: Theory and Practice. Morgan Kaufmann.

Sutton, R., Barto, A., 1998. Reinforcement Learning: An Introduction. The MIT Press.

Tambe, M., 1997. Towards flexible teamwork. Journal of Artificial Intelligence Research 7, 83–124.

Yue, Y., Gao, Y., Chapelle, O., Zhang, Y., Joachims, T., 2010. Learning more powerful test statistics for click-based retrieval evaluation. In: Proceedings of the 33rd International ACM SIGIR Conference on Research and Development in Information Retrieval. pp. 507–514.